%% file: paper.tex
\documentclass[11pt]{article}

\PassOptionsToPackage{table,dvipsnames}{xcolor}
\usepackage[preprint]{acl}

\usepackage{times}
\usepackage{latexsym}
\usepackage[T1]{fontenc}
\usepackage[utf8]{inputenc}
\usepackage{microtype}
\usepackage{inconsolata}
\usepackage{graphicx}
\usepackage{caption}
\usepackage{subcaption}
\usepackage{booktabs}
\usepackage{multirow}
\usepackage{makecell}
\usepackage{longtable}
\usepackage{colortbl}    
\usepackage{fancyvrb}
\usepackage{dblfloatfix}
\newcommand{\scoreentry}[2]{#1 & \textcolor{black!60}{#2}}
\definecolor{AnsGreen}{HTML}{1B5E20}
\definecolor{AnsRed}{HTML}{B71C1C}

\DefineVerbatimEnvironment{qualgood}{Verbatim}{%
  fontsize=\scriptsize,%
  formatcom=\fontfamily{lmtt}\selectfont,%
  frame=leftline,%
  framerule=2pt,%
  rulecolor=\color{AnsGreen},%
  xleftmargin=10pt,%
  framesep=4pt,%
  baselinestretch=1.0,%
}
\DefineVerbatimEnvironment{qualbadenv}{Verbatim}{%
  fontsize=\scriptsize,%
  formatcom=\fontfamily{lmtt}\selectfont,%
  frame=leftline,%
  framerule=2pt,%
  rulecolor=\color{AnsRed},%
  xleftmargin=10pt,%
  framesep=4pt,%
  baselinestretch=1.0,%
}
\newcommand{\qualok}{\textcolor{AnsGreen}{$\checkmark$}}
\newcommand{\qualbad}{\textcolor{AnsRed}{$\times$}}
\definecolor{QualHeaderBG}{HTML}{EEF2F7}   
\definecolor{QualMethodBG}{HTML}{FFFFFF}   
\definecolor{QualMeta}{HTML}{6B7280}       
\definecolor{QualRule}{HTML}{C9CED6}       
\newcommand{\qualmeta}[1]{{\scriptsize\color{QualMeta}#1}}
\usepackage{float}
\usepackage{placeins}
\usepackage{hyperref}
\usepackage{cuted}
\usepackage{caption}
\usepackage{placeins}
\usepackage{amsmath}
\usepackage{amssymb}
\usepackage{amsthm}
\usepackage{mathtools}
\usepackage[capitalize,noabbrev]{cleveref}
\usepackage{todonotes}

\usepackage{algorithm}
\usepackage{algorithmic}

\theoremstyle{plain}

\theoremstyle{definition}

\DeclareMathOperator*{\argmax}{arg\,max}

\newcommand{\KL}{\mathrm{KL}}

\newcommand{\conf}{\mathrm{conf}}
\newcommand{\persist}{\mathrm{persist}}
\newcommand{\jsd}{\mathrm{JSD}}
\newcommand{\accept}{\mathrm{accept}}
\newcommand{\mask}{\langle\mathrm{mask}\rangle}

\newcommand{\V}{\mathcal{V}}
\newcommand{\M}{\mathcal{M}}

\usepackage[T1]{fontenc}
\usepackage{array}
\usepackage{tabularx}
\usepackage{makecell}
\usepackage{listings}
\usepackage{pifont}

\definecolor{QualHeaderBG}{HTML}{F2F4F8}
\definecolor{QualRule}{HTML}{C8CDD7}
\definecolor{QualOK}{HTML}{22863A}
\definecolor{QualBad}{HTML}{CB2431}

\definecolor{CodeBG}{HTML}{F6F8FA}
\definecolor{CodeFrame}{HTML}{D0D7DE}
\definecolor{CodeKeyword}{HTML}{005CC5}
\definecolor{CodeString}{HTML}{032F62}
\definecolor{CodeComment}{HTML}{6A737D}

\lstdefinestyle{qualpython}{
  language=Python,
  basicstyle=\ttfamily\scriptsize,
  keywordstyle=\color{CodeKeyword}\bfseries,
  stringstyle=\color{CodeString},
  commentstyle=\color{CodeComment}\itshape,
  backgroundcolor=\color{CodeBG},
  frame=single,
  rulecolor=\color{CodeFrame},
  framerule=0.3pt,
  showstringspaces=false,
  breaklines=true,
  keepspaces=true,
  columns=fullflexible,
  tabsize=4,
  xleftmargin=1mm,
  xrightmargin=1mm,
  aboveskip=2pt,
  belowskip=1pt,
  framexleftmargin=1mm,
  framexrightmargin=1mm,
  framextopmargin=0.5mm,
  framexbottommargin=0.5mm
}

\newcolumntype{Y}{>{\raggedright\arraybackslash}X}

\title{\textit{LESS} Is More:\\ Mutual-Stability Sampling for Diffusion Language Models}

\author{
 \textbf{Amr Mohamed\textsuperscript{1,2}\thanks{Correspondence: \texttt{amr.mohamed@mbzuai.ac.ae}}},
 \textbf{Guokan Shang\textsuperscript{1}},
 \textbf{Michalis Vazirgiannis\textsuperscript{1,2}}
\\
\\
 \textsuperscript{1}MBZUAI,
 \textsuperscript{2}Ecole Polytechnique
\\
}

\begin{document}
\maketitle

\input{sections/00_abstract}
\input{sections/01_intro}
\input{sections/02_related}
\input{sections/04_method}
\input{sections/05_experiments}
\input{sections/06_ablations}
\input{sections/07_discussion}
\input{sections/08_conclusion}
\bibliography{references}

\input{sections/09_appendix}

\end{document}

%% file: sections/00_abstract.tex
\begin{abstract}
Diffusion large language models (dLLMs) offer a promising alternative to autoregressive decoding by iteratively refining masked sequences, enabling parallel token updates and bidirectional conditioning. Their practical efficiency, however, is limited by sampling procedures that execute a fixed number of reverse denoising steps selected before decoding, spending computation on already-stable positions and sometimes committing unstable ones too early. We present \textsc{LESS}, a training-free, model-agnostic adaptive sampler that treats token commitment as an online stopping problem. \textsc{LESS} implements mutual-stability sampling through a joint stability rule that makes a masked position eligible for unmasking only when its top-1 prediction has high confidence, its top-1 token persists across recent reverse steps, and its predictive distribution is stable under top-$K$ inter-step Jensen--Shannon divergence. We evaluate \textsc{LESS} on Dream-7B, LLaDA-8B, and LLaDA-1.5-8B, covering full-sequence diffusion and semi-autoregressive blockwise sampling regimes, across seven benchmarks spanning general knowledge, math, and code. \textsc{LESS} improves average accuracy over strong training-free adaptive samplers while using $72.1\%$ fewer reverse steps than fixed-budget decoding. Since each reverse step requires a Transformer forward pass, these step-count reductions translate into fewer forward evaluations, lower measured wall-clock latency, and lower estimated inference compute.
\end{abstract}

%% file: sections/01_intro.tex
\section{Introduction}
\label{sec:intro}

\begin{figure*}[!htbp]
    \centering
    \includegraphics[width=\textwidth]{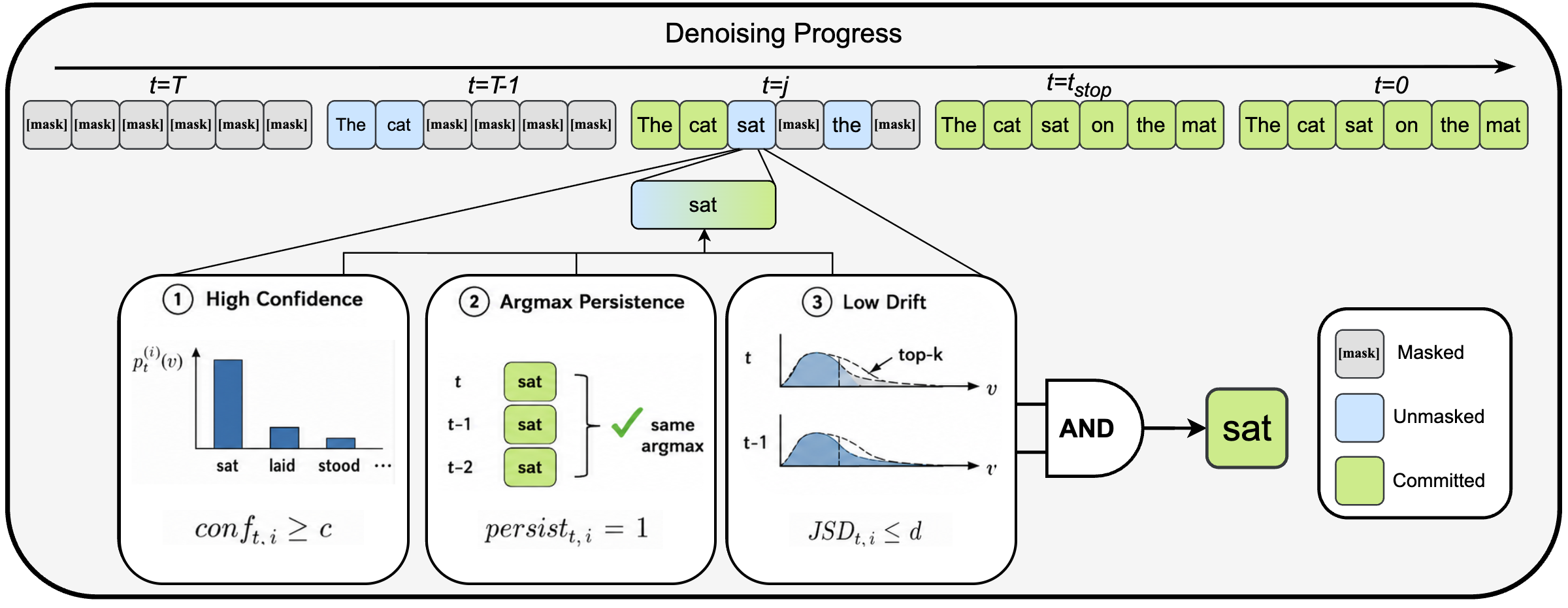}
    \caption{\textbf{Overview of \textsc{LESS}.} Masked diffusion decoding iteratively unmasks a fully masked sequence. For each masked position $i$, \textsc{LESS} computes three stability signals on the predictive distribution $p_{t,i}$: top-1 predictive confidence, top-1 token persistence, and top-$K$ inter-step Jensen--Shannon divergence between consecutive predictive distributions. A position is eligible under the joint stability rule only when all conditions are met.}
    \label{fig:illustration}
\end{figure*}
Large language models (LLMs) have advanced rapidly, yet text generation remains dominated by autoregressive (AR) decoding, limiting opportunities for global refinement \citep{brown2020language,yin2024survey,zhang2025survey}. Diffusion large language models (dLLMs) have emerged as a compelling alternative: instead of generating one token at a time, they iteratively denoise masked sequences, enabling parallel token updates, bidirectional conditioning over partially generated text, and natural infilling \citep{Zou2023-gr}. This paradigm has matured quickly, with recent work training capable diffusion LMs from scratch \citep{Nie2025-ii,Ye2025-hm,Yang2025-vrpo}, adapting AR checkpoints to diffusion objectives \citep{Gong2024-on}, and extending dLLMs to reasoning, planning, and code-generation \citep{d1_2025,BeyondAR_2024,Song2025-qp,MercuryCoder_2025}. Together with scaling evidence \citep{ScalingMDM_2024}, these advances position dLLMs as practical language modeling architectures.

Despite this promise, decoding efficiency remains a central bottleneck for dLLMs. Generation proceeds through a reverse-diffusion chain with many refinement steps, and standard samplers require the reverse-step budget and unmasking schedule to be chosen \emph{a priori}, often conservatively to avoid quality loss across tasks \citep{Nie2025-ii,Ye2025-hm,Yang2025-vrpo}. This fixed schedule ignores position-wise convergence: empirical studies of adaptive dLLM decoding show that some token predictions stabilize well before the final reverse step \citep{Pengxiang2025-eq,kim2025klass,mohamed2025fast}. As a result, some positions continue to be recomputed after their local predictive distributions have stabilized, while others may be unmasked before those distributions have settled. Since most dLLM samplers treat unmasked tokens as fixed, premature commitment is difficult to recover from: once an incorrect token is written, later reverse steps typically cannot revise it.

A growing body of work addresses this bottleneck from complementary directions. Training-time methods shorten the reverse process through distillation, curricula, or flow matching \citep{Sahoo2025-vf,Chen2025-lj,FSDFM_2024}, while systems methods reduce per-step latency through caching or speculative execution \citep{dKVCache_2025,d2Cache_2025,Spiffy_2025,SelfSpeculative_2025}. Training-free samplers instead adapt when positions are unmasked, using confidence, margin, or inter-step divergence signals from the model's predictive distributions \citep{Pengxiang2025-eq,kim2025klass,wu2025fastdllm,mohamed2025fast}. These methods show that predictions often stabilize before the final diffusion step, but leave open a fundamental question: \emph{when is a masked position stable enough to stop refining and commit?}

We address this question by framing adaptive unmasking as per-position online stopping. Prior training-free samplers provide partial criteria: confidence- and margin-based methods test whether the current prediction is locally decisive, while divergence-aware methods test whether consecutive predictive distributions are close \citep{Pengxiang2025-eq,wu2025fastdllm,mohamed2025fast,kim2025klass}. \textsc{LESS} extends this line by treating stability as a joint property of the current prediction and its recent reverse-time trajectory. It implements mutual-stability sampling with a joint rule: a position is eligible for unmasking only when its top-1 prediction is confident, its top-1 token persists across recent reverse steps, and its top-$K$ inter-step Jensen--Shannon divergence is small.

The top-$K$ inter-step Jensen--Shannon divergence (JSD) is the key distributional signal in \textsc{LESS}. It provides a bounded, symmetric measure of how much a position's predictive distribution changes between reverse steps. This matters as confidence alone cannot reveal ongoing distributional motion, and a fixed top-1 token can hide redistribution among plausible alternatives. Together, the three signals yield a training-free, model-agnostic joint stability rule computed from predictive distributions already produced during decoding.

We evaluate \textsc{LESS} on three dLLMs across seven math, code, and general-knowledge benchmarks. \textsc{LESS} improves the accuracy--efficiency trade-off on average over strong training-free adaptive samplers, with the most pronounced gains on math and code. Ablations show that confidence is the primary acceptance signal, while persistence and inter-step JSD provide complementary safeguards. FLOPs profiling and wall-clock measurements confirm that step reductions translate into fewer Transformer forward evaluations and lower decoding latency. Our contributions are:\vspace{-4pt}
\begin{itemize}
    \item We formulate dLLM token commitment as a per-position online stopping problem induced by irreversible unmasking.\vspace{-4pt}
    \item We introduce top-$K$ inter-step Jensen--Shannon divergence as a bounded, symmetric stability signal for dLLM decoding.\vspace{-4pt}
    \item We present \textsc{LESS}, a training-free sampler that combines confidence, top-1 token persistence, and inter-step JSD in a joint stability rule.\vspace{-4pt}
    \item We validate \textsc{LESS} across three dLLM families and seven benchmarks, showing improved average accuracy--efficiency trade-offs with measured compute savings.\vspace{-4pt}
\end{itemize}
\textsc{LESS} reduces reverse denoising steps through training-free, stability-gated unmasking. Our code is publicly available.\footnote{https://github.com/amr-mohamedd/LESS-Is-More}

%% file: sections/02_related.tex
\section{Related Work}
\label{sec:related}

\paragraph{Masked diffusion language models.}
Masked diffusion language models generate text by iteratively denoising partially masked sequences, exposing per-position predictive distributions at each reverse step. Recent dLLMs such as LLaDA, Dream, and LLaDA-1.5 support instruction following, reasoning, and code generation \citep{Nie2025-ii,Ye2025-hm,Yang2025-vrpo}. Related works study adaptation from AR checkpoints, scaling behavior, long-context extension, and code-specialized variants \citep{Gong2024-on,ScalingMDM_2024,UltraLLaDA_2025,Song2025-qp}. These efforts position dLLMs as a practical alternative to AR generation across diverse generation settings, but inference remains dominated by reverse denoising steps.

\paragraph{Efficient dLLM inference.}
Prior acceleration methods reduce this cost in three complementary ways. Training-time methods shorten the reverse process through curricula, consistency-style distillation, or flow matching \citep{Sahoo2025-vf,Chen2025-lj,FSDFM_2024}; systems methods reduce per-step cost through diffusion-aware caching, speculative decoding, or related reuse mechanisms \citep{dKVCache_2025,d2Cache_2025,Spiffy_2025,SelfSpeculative_2025}; and inference-time adaptive samplers change \emph{which} positions are unmasked and \emph{when}. \textsc{LESS} belongs to the third class and is orthogonal to the first two: it reduces the number of reverse steps and can in principle be combined with faster per-step execution.

\paragraph{Adaptive token commitment.}
Training-free adaptive samplers typically accelerate dLLM decoding by committing tokens once a local reliability criterion is satisfied. Confidence-based rules unmask positions whose top prediction or top-1/top-2 margin is large \citep{Pengxiang2025-eq,wu2025fastdllm,mohamed2025fast}; these methods are effective when confidence is well aligned with convergence, but confidence alone does not explicitly track recent changes in the predicted token or continued motion of the predictive distribution. Divergence-aware methods address this limitation by adding inter-step stability checks, such as token-level KL between consecutive per-position predictive distributions \citep{kim2025klass}, although KL-based criteria are directional and unbounded. \textsc{LESS} extends this line by treating adaptive unmasking as per-position online stopping: a position is accepted only when confidence, top-1 token persistence, and inter-step distributional stability agree. In particular, \textsc{LESS} uses a bounded, symmetric top-$K$ JSD criterion and adds an explicit persistence gate for the predicted token identity. To our knowledge, \textsc{LESS} is the first training-free dLLM sampler to use JSD as an inter-step stability signal for adaptive token commitment.

%% file: sections/04_method.tex
\section{Methods}
\label{sec:method}
\label{sec:bg}

We first define the masked-diffusion decoding setup, then present \textsc{LESS}, a training-free adaptive sampler that treats token-unmasking as a per-position online stopping problem. \textsc{LESS} implements mutual-stability sampling through a joint stability rule over top-1 confidence, top-1 token persistence, and top-$K$ inter-step JSD.

\subsection{Preliminaries: masked diffusion language models}
\label{sec:method:prelim}

\paragraph{Setup.}
Let $\V$ be the vocabulary and let $\mask$ denote the mask token. Given a prompt $x_{\mathrm{prompt}}$, a masked diffusion language model generates a length-$L$ continuation by denoising a sequence
\[
x_t \in (\V \cup \{\mask\})^L,
\quad t = T,T-1,\ldots,0,
\]
where $x_T$ is fully masked and $x_0$ is fully decoded. At reverse step $t$, the model observes the prompt and current partially decoded sequence $x_t$, then produces logits
\begin{equation}
L_t
=
f_\theta(x_{\mathrm{prompt}}, x_t, t)
\in \mathbb{R}^{L \times |\V|},
\label{eq:logits}
\end{equation}
with per-position predictive distributions
\begin{equation}
p_{t,i}(\cdot)
=
\mathrm{softmax}(L_{t,i})
\in \Delta^{|\V|-1}.
\label{eq:model-marginal}
\end{equation}
We write
\begin{equation}
\M_t \coloneqq \{i : x_{t,i}=\mask\}
\end{equation}
for the still-masked positions.

\paragraph{Forward masking process.}
During training, clean sequences $x_0 \in \V^L$ are corrupted by progressively replacing tokens with $\mask$. With step-dependent masking rate $\beta_t \in [0,1)$, the forward transition is
\begin{equation}
\!
\begin{aligned}
q(x_t\mid x_{t-1})
&= \prod_{i=1}^{L}\Big[(1-\beta_t)\delta(x_{t,i}=x_{t-1,i})
\\&\hspace{2.4em}
+ \beta_t\,\delta(x_{t,i}=[\mask])\Big].
\end{aligned}
\label{eq:q_step}
\end{equation}
Writing $\bar\alpha_t = \prod_{s=1}^{t}(1-\beta_s)$ for the token survival probability after $t$ steps,
the $t$-step marginal becomes
\begin{equation}
\!
\begin{aligned}
q(x_t\mid x_0,t)
&= \prod_{i=1}^{L}\Big[\bar\alpha_t\,\delta(x_{t,i}=x_{0,i})
\\&\hspace{-0.1em}
+ (1-\bar\alpha_t)\,\delta(x_{t,i}=[\mask])\Big].
\end{aligned}
\label{eq:q_closed}
\end{equation}
The mask token is absorbing in the forward process.

\paragraph{Reverse denoising process.}
The learned reverse process provides clean-token predictions for positions that remain masked. For each $i\in\mathcal M_t$, the model predictive distribution in Eq.~\ref{eq:model-marginal} parameterizes
\begin{equation}
\!
\begin{aligned}
p_\theta(\widehat{x}_{t,i}\mid x_{\mathrm{prompt}},x_t)
=\\
&\hspace{-5em}\mathrm{Cat}\!\left(\widehat{x}_{t,i};p_{t,i}(\cdot\mid x_{\mathrm{prompt}},x_t)\right).
\end{aligned}
\end{equation}
Positions that are already unmasked are carried forward deterministically:
\begin{equation}
\!
\begin{aligned}
p_\theta(x_{t-1,i}\mid x_{\mathrm{prompt}},x_t)
=\\
&\hspace{-5em}\delta(x_{t-1,i}=x_{t,i}), \quad i\notin\mathcal M_t .
\end{aligned}
\end{equation}
Thus, stochastic or argmax token selection is applied only to positions chosen for unmasking; all other positions are copied unchanged.

\paragraph{Partial unmasking.}
In practice, a reverse step usually unmasks only a subset of masked positions. Let
\[
\mathcal U_t \subseteq \M_t
\]
be the unmasking set at step $t$. The update is
\begin{equation}
x_{t-1,i}
=
\begin{cases}
x_{t,i}, & i \notin \mathcal U_t,\\
\widehat x_{t,i}, & i \in \mathcal U_t,
\end{cases}
\label{eq:step}
\end{equation}
where $\widehat x_{t,i}$ is selected from $p_{t,i}$, either by argmax or by the model's configured stochastic decoding rule. Because unmasked tokens are typically carried over in subsequent reverse steps, each unmasking decision can be viewed as a token commitment. Standard samplers choose $\mathcal U_t$ using a fixed unmasking schedule, for example by unmasking a predetermined number of high-confidence positions at each step. This is the decision \textsc{LESS} replaces: rather than unmasking according to a static budget, \textsc{LESS} asks whether each position is stable enough to stop.

\paragraph{Training objective.}
Masked diffusion language models are trained to reconstruct clean tokens at masked positions. With timesteps sampled uniformly, the standard masked-token objective is
\begin{equation}
\!
\begin{aligned}
\mathcal{L}(\theta)
&= \mathbb{E}_{x_0\sim\mathcal{D}}
   \mathbb{E}_{t\sim\mathcal{U}\{1{:}T\}}
   \mathbb{E}_{x_t\sim q(\cdot\mid x_0,t)}
\\&
\left[-\sum_{i\in \M_t}
\log p_{t,i}\big(x_{0,i} \mid x_{\mathrm{prompt}},x_t\big)\right].
\end{aligned}
\label{eq:loss}
\end{equation}
This objective yields the per-position predictive distributions used by the reverse sampler, but it does not prescribe when a predicted token should be committed during inference.

\subsection{Adaptive unmasking as online stopping}
\label{sec:method:stopping}

Because committed tokens are typically not revised, selecting $\mathcal U_t$ is an irreversible token-commitment decision. LESS therefore treats adaptive unmasking as a per-position online stopping problem. For each position $i$, the sampler observes a trajectory of predictive distributions over reverse steps and chooses a stopping time that is adapted to the information available during decoding. Once $i$ is unmasked, its selected token is carried forward by the reverse process and is not revised by later steps. The goal is to stop early enough to avoid recomputing positions whose local predictions have already stabilized, while avoiding premature commitments before the predictive trajectory has settled. LESS implements this stopping rule through three observable stability signals: current top-1 confidence, persistence of the top-1 token over recent observations, and inter-step distributional stability under top-$K$ Jensen--Shannon divergence.

\subsection{Stability signals}
\label{sec:method:signals}

For each still-masked position $i\in\M_t$, define the top-1 token
\begin{equation}
w_{t,i}
\coloneqq
\argmax_{v \in \V} p_{t,i}(v),
\end{equation}

\paragraph{Top-1 predictive confidence.}
The confidence signal is the probability assigned to the current top-1 token:
\begin{equation}
\conf_{t,i}
\coloneqq
p_{t,i}(w_{t,i})
=
\max_{v \in \V} p_{t,i}(v).
\label{eq:conf}
\end{equation}

\paragraph{Top-1 token persistence.}
High confidence alone does not guarantee temporal stability: the top-1 token may have just changed. \textsc{LESS} therefore requires the current top-1 token to remain unchanged over a persistence window of length $P$:
\begin{equation}
\persist_{t,i}
\coloneqq
\prod_{k=1}^{P-1}
\mathbf{1}\!\left[
w_{t+k,i}=w_{t,i}
\right].
\label{eq:persist}
\end{equation}
For steps with insufficient history, we set $\persist_{t,i}=0$.

\paragraph{Top-$K$ inter-step Jensen--Shannon divergence.}
The third signal measures whether the local predictive distribution itself has stopped changing across reverse steps. For a masked position $i$, \textsc{LESS} compares $p_{t,i}$ and $p_{t+1,i}$ using a Jensen--Shannon divergence on a coarsened vocabulary. Let
\begin{equation}
A_{t,i}
\coloneqq
\mathrm{top}\text{-}K(p_{t,i})
\cup
\mathrm{top}\text{-}K(p_{t+1,i})
\label{eq:union-support}
\end{equation}
be the union of the consecutive top-$K$ candidate sets. All remaining vocabulary mass is collapsed into a residual symbol $\bot$. Define
\begin{equation}
\phi_{t,i}(v)
=
\begin{cases}
v, & v \in A_{t,i},\\
\bot, & v \notin A_{t,i},
\end{cases}
\label{eq:projection}
\end{equation}
and let
\begin{gather}
\widetilde p_{t,i}
\coloneqq
(\phi_{t,i})_{\#}p_{t,i},\\
\widetilde p_{t+1,i}
\coloneqq
(\phi_{t,i})_{\#}p_{t+1,i}
\label{eq:coarsened}
\end{gather}
be the coarsened predictive distributions on $A_{t,i}\cup\{\bot\}$. With
\begin{equation}
m_{t,i}
\coloneqq
\tfrac{1}{2}
\left(
\widetilde p_{t,i}
+
\widetilde p_{t+1,i}
\right),
\end{equation}
the inter-step JSD is
\begin{equation}
\!
\begin{aligned}
\jsd_{t,i}&\coloneqq\tfrac{1}{2}\KL(\widetilde p_{t,i}\,\|\,m_{t,i})
\\&+\tfrac{1}{2}\KL(\widetilde p_{t+1,i}\,\|\,m_{t,i}).
\end{aligned}
\label{eq:drift}
\end{equation}
At the first reverse observation, where no previous predictive distribution is available, we set $\jsd_{t,i}=+\infty$. The signal tests whether the token-level belief at position $i$ has become locally stationary under ongoing context updates. Since $p_{t,i}$ and $p_{t+1,i}$ are produced under different partially unmasked contexts, the comparison should not depend on which step is treated as the reference. One-way divergences such as $\KL(p_{t,i}\,\|\,p_{t+1,i})$ are less suitable for this inter-step stability test because KL divergence is asymmetric and unbounded \citep{CRMATH_2025__363_G8_809_0,shone2025beyond,qiu2025multimodal}. By contrast, JSD is symmetric and bounded:
\begin{equation}
0 \le \jsd_{t,i} \le \log 2
\end{equation}
under natural logarithms \citep{e22020221,struckman2025should}. The bounded scale makes the threshold $d$ easier to share across positions, timesteps, models, and tasks. The top-$K$ projection focuses the comparison on tokens most likely to affect the top-1 decision, while the residual bin preserves total mass.
\subsection{Joint stability rule}
\label{sec:method:rule}

A masked position is eligible under the joint stability rule if and only if all three stability tests pass:
\begin{equation}
\boxed{
\begin{aligned}
\accept_{t,i}
\;=\;&
\bigl(\conf_{t,i} \ge c\bigr)
\,\wedge\,
\persist_{t,i}\quad
\\
&{}\wedge\,
\bigl(\jsd_{t,i} \le d\bigr).
\end{aligned}
}
\label{eq:accept}
\end{equation}
where $c$ is the confidence threshold and $d$ is the inter-step JSD threshold; larger $c$ and smaller $d$ make the rule more conservative. The conjunction is intentional: the three tests respectively check decisiveness, top-1 temporal stability, and distributional stationarity. \textsc{LESS} accepts a position only when all three agree.

\subsection{Unmasking policy}
\label{sec:method:commit}

The joint stability rule identifies eligible positions; the unmasking policy chooses among them. For full-sequence diffusion samplers, where positions may be selected across the full-generation span, \textsc{LESS} uses a frontier-first policy. Let
\begin{equation}
f_t \coloneqq \min \M_t
\end{equation}
be the leftmost still-masked position. If $f_t$ is eligible, \textsc{LESS} unmasks it; otherwise, it scans a look-ahead window of size $W$ and unmasks the first eligible position:
\begin{equation}
\begin{aligned}
j^\star
=
\min
\{j : f_t < j \le f_t+W,\; j \in \M_t,\;\\ \accept_{t,j}=1\}.
\end{aligned}
\end{equation}
If no eligible position is found, the sampler skips the step. After $\Pi$ consecutive skipped steps, \textsc{LESS} performs a forced-unmasking step using the underlying sampler's default action. This fallback guarantees termination but is not counted as accepted by the joint stability rule. For semi-autoregressive blockwise samplers, \textsc{LESS} applies the same eligibility rule inside the active block and preserves the model's blockwise decoding interface.


%% file: sections/05_experiments.tex
\begin{table*}[t]
\centering
\scriptsize
\setlength{\tabcolsep}{2.5pt}
\renewcommand{\arraystretch}{1.1}
\resizebox{\textwidth}{!}{%
\begin{tabular}{l *{8}{cc}}
\toprule
& \multicolumn{4}{c}{\emph{Math}} & \multicolumn{4}{c}{\emph{Code}} & \multicolumn{6}{c}{\emph{General Knowledge}} & \multicolumn{2}{c}{} \\
\cmidrule(lr){2-5} \cmidrule(lr){6-9} \cmidrule(lr){10-15}
& \multicolumn{2}{c}{GSM8K} & \multicolumn{2}{c}{MATH} & \multicolumn{2}{c}{HumanEval} & \multicolumn{2}{c}{MBPP} & \multicolumn{2}{c}{MMLU} & \multicolumn{2}{c}{HellaSwag} & \multicolumn{2}{c}{WinoGrande} & \multicolumn{2}{c}{Mean} \\
\cmidrule(lr){2-3} \cmidrule(lr){4-5} \cmidrule(lr){6-7} \cmidrule(lr){8-9} \cmidrule(lr){10-11} \cmidrule(lr){12-13} \cmidrule(lr){14-15} \cmidrule(lr){16-17}
Method & Acc & Steps & Acc & Steps & Acc & Steps & Acc & Steps & Acc & Steps & Acc & Steps & Acc & Steps & Acc & Steps \\
\midrule
\multicolumn{17}{c}{\underline{\textbf{\emph{Dream-7B}}}} \vspace{3pt}\\
Base          & \scoreentry{79.61}{256.0}          & \scoreentry{37.80}{256.0}          & \scoreentry{57.93}{256.0}         & \scoreentry{61.87}{256.0}         & \scoreentry{72.32}{256.0}         & \scoreentry{69.72}{256.0}         & \scoreentry{71.82}{256.0}         & \scoreentry{63.30}{256.0} \\
Prophet       & \scoreentry{48.22}{170.2}          & \scoreentry{20.00}{174.8}          & \scoreentry{51.22}{94.3}          & \scoreentry{57.20}{111.5}         & \scoreentry{70.82}{2.2}           & \scoreentry{70.77}{4.6}           & \scoreentry{\textbf{72.22}}{3.1}  & \scoreentry{53.20}{92.9} \\
KLASS         & \scoreentry{79.61}{156.2}          & \scoreentry{41.40}{147.7}          & \scoreentry{58.54}{74.0}          & \scoreentry{62.26}{111.7}         & \scoreentry{\textbf{72.02}}{5.6}  & \scoreentry{70.49}{20.2}          & \scoreentry{66.30}{8.0}           & \scoreentry{63.50}{85.4} \\
\rowcolor{gray!30}
\textbf{\textsc{LESS} (ours)} & \scoreentry{\textbf{81.88}}{106.5} & \scoreentry{\textbf{41.60}}{122.5} & \scoreentry{\textbf{60.98}}{48.8} & \scoreentry{\textbf{63.42}}{98.3} & \scoreentry{70.93}{2.2}           & \scoreentry{\textbf{72.48}}{8.3}  & \scoreentry{71.43}{4.0}           & \scoreentry{\textbf{65.18}}{64.3} \\
\midrule
\multicolumn{17}{c}{\underline{\textbf{\emph{LLaDA-8B}}}} \vspace{3pt}\\
Base          & \scoreentry{75.44}{256.0}          & \scoreentry{31.40}{256.0}          & \scoreentry{37.20}{256.0}         & \scoreentry{48.25}{256.0}         & \scoreentry{65.63}{256.0}         & \scoreentry{78.13}{256.0}         & \scoreentry{77.03}{256.0}         & \scoreentry{56.58}{256.0} \\
Prophet       & \scoreentry{37.68}{176.2}          & \scoreentry{14.20}{176.2}          & \scoreentry{25.61}{180.9}         & \scoreentry{44.36}{180.3}         & \scoreentry{64.44}{1.2}           & \scoreentry{78.06}{78.5}          & \scoreentry{75.69}{3.1}           & \scoreentry{44.55}{128.1} \\
KLASS         & \scoreentry{75.13}{98.7}           & \scoreentry{32.60}{126.1}          & \scoreentry{37.80}{93.6}          & \scoreentry{49.42}{120.3}         & \scoreentry{\textbf{65.63}}{11.7} & \scoreentry{78.12}{80.4}          & \scoreentry{77.03}{31.1}          & \scoreentry{57.02}{86.8} \\
\rowcolor{gray!30}
\textbf{\textsc{LESS} (ours)} & \scoreentry{\textbf{75.82}}{82.0}  & \scoreentry{\textbf{33.00}}{125.6} & \scoreentry{\textbf{38.41}}{71.7} & \scoreentry{\textbf{49.81}}{115.5} & \scoreentry{65.61}{12.6}          & \scoreentry{\textbf{78.21}}{65.4} & \scoreentry{\textbf{77.35}}{18.4} & \scoreentry{\textbf{57.41}}{76.5} \\
\midrule
\multicolumn{17}{c}{\underline{\textbf{\emph{LLaDA-1.5-8B}}}} \vspace{3pt}\\
Base          & \scoreentry{77.33}{256.0}          & \scoreentry{34.00}{256.0}          & \scoreentry{40.85}{256.0}         & \scoreentry{47.47}{256.0}         & \scoreentry{65.89}{256.0}         & \scoreentry{77.86}{256.0}         & \scoreentry{77.35}{256.0}         & \scoreentry{57.84}{256.0} \\
Prophet       & \scoreentry{38.97}{172.5}          & \scoreentry{14.20}{169.8}          & \scoreentry{27.44}{173.9}         & \scoreentry{43.19}{173.3}         & \scoreentry{64.58}{1.2}           & \scoreentry{77.86}{93.7}          & \scoreentry{75.69}{5.1}           & \scoreentry{44.87}{126.0} \\
KLASS         & \scoreentry{74.00}{98.4}           & \scoreentry{33.80}{127.8}          & \scoreentry{\textbf{41.46}}{101.3} & \scoreentry{47.47}{123.6}        & \scoreentry{65.89}{11.7}          & \scoreentry{77.86}{101.2}         & \scoreentry{77.35}{34.5}          & \scoreentry{57.36}{91.6} \\
\rowcolor{gray!30}
\textbf{\textsc{LESS} (ours)} & \scoreentry{\textbf{76.12}}{86.3}  & \scoreentry{\textbf{34.80}}{107.5} & \scoreentry{40.85}{69.6}          & \scoreentry{\textbf{47.86}}{97.2} & \scoreentry{\textbf{65.91}}{12.6} & \scoreentry{\textbf{77.88}}{84.5} & \scoreentry{\textbf{77.43}}{21.2} & \scoreentry{\textbf{57.85}}{73.2} \\
\bottomrule
\end{tabular}}
\caption{Per-benchmark comparison across three dLLMs and seven benchmarks. Each entry reports accuracy and average executed reverse steps per sample under maximum reverse-step budget $T_{\max}=256$. Base is the fixed-budget reference and executes the full budget. Adaptive samplers may terminate earlier once the task-specific completion condition is met. Bold marks the best accuracy among training-free adaptive samplers for each benchmark and model. The Mean column reports a category-macro average over math, code, and general-knowledge tasks.}
\label{tab:headline}
\label{tab:headline}
\end{table*}

\section{Experimental Setup}
\label{sec:exp}

\paragraph{Models.} We evaluate instruction-tuned checkpoints from three masked dLLM families: Dream-7B \citep{Ye2025-hm}, LLaDA-8B \citep{Nie2025-ii}, and LLaDA-1.5-8B \citep{Yang2025-vrpo}. These checkpoints cover two reverse-sampling regimes. Dream-7B uses full-sequence diffusion sampling, allowing positions to be unmasked anywhere in the generation span. LLaDA-8B and LLaDA-1.5-8B use semi-autoregressive blockwise sampling, where unmasking is restricted to the active block.

\paragraph{Benchmarks.} Evaluation spans seven benchmarks across three task categories. \emph{Math and reasoning:} GSM8K \citep{cobbe2021gsm8k} and MATH \citep{hendrycks2021math}. \emph{Code generation:} HumanEval \citep{chen2021humaneval} and MBPP \citep{austin2021mbpp}. \emph{General knowledge and commonsense:} MMLU \citep{hendrycks2020mmlu}, HellaSwag \citep{zellers2019hellaswag}, and WinoGrande \citep{sakaguchi2021winogrande}. Additional evaluation details, including generation lengths and reverse-step budgets, are provided in Appendix~\ref{app:eval-protocol}.

\paragraph{Baselines.} We compare \textsc{LESS} against three classes of decoding strategies. \textit{Base} denotes the canonical released fixed-budget sampler for each model. \textit{Top-2 confidence-gap early-commit decoding} is represented by Prophet \citep{Pengxiang2025-eq}, which uses the gap between the top-2 prediction candidates to decide when to stop refinement and decode all remaining tokens in one step. \textit{Token-level divergence-aware unmasking} is represented by KLASS \citep{kim2025klass}, which uses token-level KL divergence together with confidence to select stable, high-confidence tokens for early unmasking. All baselines are evaluated using their released decoding configurations.

\paragraph{Configuration.} Unless otherwise stated, all results use the fixed global \textsc{LESS} configuration $(c,d)=(0.75,0.040)$ across models and benchmarks. The remaining parameters are frozen across all experiments: top-$K$ size $K=8$, persistence length $P=2$, look-ahead window $W=16$, and skip budget $\Pi=2$.

\paragraph{Metrics.}
We report task accuracy and the average number of reverse denoising steps executed per sample. We additionally report FLOPs profiling and wall-clock measurements in Section~\ref{subsec:flops}.
\section{Results}
\label{sec:results}

Table~\ref{tab:headline} summarizes the main results across all three model families and seven benchmarks. \textsc{LESS} improves the aggregate accuracy--efficiency trade-off over strong training-free adaptive samplers.

Across the three model families, \textsc{LESS} improves or matches the category-macro mean relative to Base while using substantially fewer denoising steps. On Dream-7B, \textsc{LESS} reaches $65.18$ mean accuracy ($\Delta{+}1.88$) with $64.3$ steps. On LLaDA-8B, it reaches $57.41$ mean accuracy ($\Delta{+}0.83$) with $76.5$ steps. On LLaDA-1.5-8B, it achieves comparable mean accuracy, $57.85$ ($\Delta{+}0.01$), with $73.2$ steps. Averaged across the three model-family means, \textsc{LESS} uses $72.1\%$ fewer reverse steps than Base while yielding average accuracy gains of $\Delta{+}0.91$ points. \textsc{LESS} also outperforms both baselines on average for every model family, achieving higher mean accuracy with fewer mean steps than both Prophet and KLASS. These gains are achieved with a fixed global configuration, indicating that the stability signals used by \textsc{LESS} transfer consistently across tasks, models, and decoding regimes.

%% file: sections/06_ablations.tex
\section{Ablations and Analysis}
\label{sec:abl}

We analyze \textsc{LESS} along four axes: the contribution of each stability signal to robust unmasking, the extent to which decoding is governed by rule-accepted unmaskings rather than the forced-unmasking fallback, the role of the unmasking policy across decoding regimes, and whether reductions in reverse steps translate into compute savings. Full ablation results are provided in Appendix~\ref{app:ablations}.

\subsection{Complementarity of the stability signals}
\label{sec:abl:signals}

We test whether the three signals in the joint stability rule provide complementary safeguards by comparing full \textsc{LESS} with single-signal and drop-one variants on Dream-7B and LLaDA-8B. Under the fixed global-threshold setting, no individual signal provides the same reliable operating point as the full rule. Confidence is the strongest standalone signal on average, while JSD-only, persistence-only, and margin-only exhibit substantially weaker aggregate performance across model families and benchmarks (Appendix Table~\ref{tab:abl-single}). This indicates that JSD and persistence are not intended to act as standalone unmasking criteria; rather, they serve as stabilizing gates when combined with confidence.

Drop-one variants show the same pattern in reverse. Removing confidence causes the largest degradation, confirming its role as the primary acceptance gate. Removing inter-step JSD or persistence is less damaging, but the resulting two-signal variants are less stable across models and benchmarks than the full rule (Appendix Table~\ref{tab:abl-pair}). These results support the conjunctive design of \textsc{LESS}: confidence provides decisiveness, persistence guards against recent top-1 changes, and JSD guards against continuing distributional motion.

\subsection{Rule-driven versus forced unmasking}
\label{sec:abl:fallback}

The forced-unmasking fallback guarantees progress when no position passes the joint stability rule by forcing at least one required position to be unmasked after $\Pi$ consecutive skipped steps. In the full-sequence Dream-7B setting, where the rule can directly govern unmasking decisions, the fallback is invoked on only $2$--$9\%$ of reverse steps across audited benchmarks. Thus, most Dream-7B reverse steps are governed directly by the joint stability rule rather than by fallback progress. We also audit top-1 token changes over reverse time. As shown for Dream-7B in Figure 2, the top-1 flip rate is highest early and decays steadily, reaching at most $5\%$ in the final quarter of the executed trajectories. This supports the temporal-stability assumption underlying \textsc{LESS}: per-position predictive distributions tend to stabilize over reverse time, but early commitments can occur before that stabilization is complete. Further details are provided in Appendix~\ref{app:abl:fallback}.

\begin{figure*}[t]
    \centering
    \includegraphics[width=\textwidth]{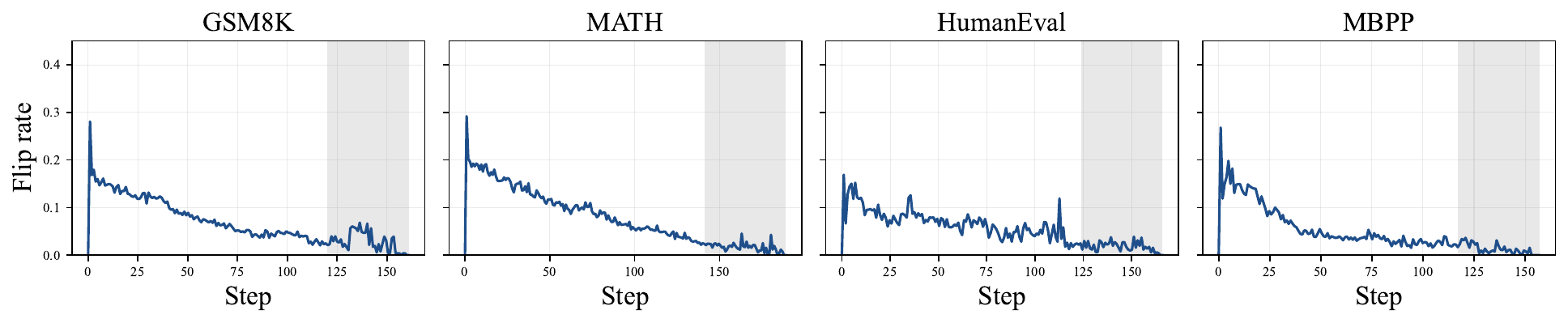}
    \caption{Dream-7B per-step top-1 token flip rate over generation positions ($T=256$) averaged over 50 samples per benchmark. The shaded region marks the final 25\% of reverse steps.}
    \label{fig:flip-rate-dream}
\end{figure*}

\subsection{Robustness across decoding regimes}
\label{sec:abl:scope}

To isolate the effect of the unmasking policy, we compare the frontier-first policy with a parallel-unmasking variant that commits all eligible positions at each reverse step. On Dream-7B, parallel unmasking lowers accuracy on three of four math/code benchmarks, with an average change of $-2.7$ accuracy points and little aggregate change in step count. This is consistent with the interpretation that committing many eligible positions simultaneously can perturb subsequent denoising states before local context has fully settled. On LLaDA-8B, accuracy is unchanged because its released semi-autoregressive blockwise sampler already commits accepted positions in parallel within the active block. Additional budget and token-selection ablations show that $T=256$ reaches the rule's operating point for the evaluated settings: increasing to $T=512$ produces no material aggregate improvement, reducing to $T=128$ degrades harder settings, and nucleus sampling affects accuracy without changing the eligibility criterion at each reverse step. Further details are provided in Appendix~\ref{app:ablations}.

\subsection{Step reductions and compute savings}
\label{subsec:flops}

Each reverse step requires a Transformer forward pass; thus, fewer denoising steps reduce inference cost. To assess this effect, we conduct FLOPs profiling on Dream-7B GSM8K under reverse-step budgets $T \in \{32,64,128,256\}$. Mean estimated per-example speedup over the fixed-budget sampler increases from $1.40\times$ at $T{=}32$ to $5.01\times$ at $T{=}256$ (Appendix~\ref{app:flops_details}); at $T{=}256$, \textsc{LESS} executes $64.5$ Transformer forward evaluations on average. 
\begin{figure}[!htbp]
    \centering
    \includegraphics[width=\columnwidth]{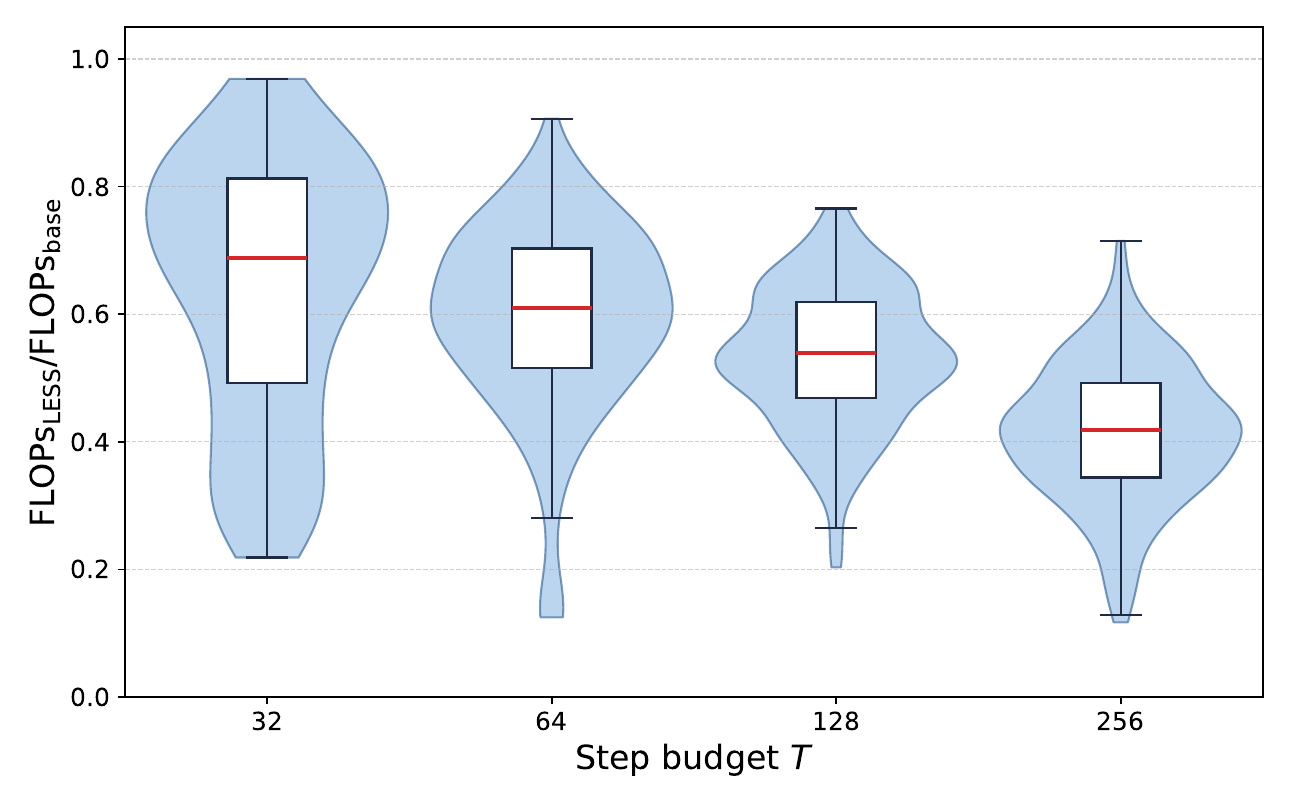}
    \caption{Per-example compute ratio $\mathrm{FLOPs}_{\textsc{LESS}}/\mathrm{FLOPs}_{\textsc{Base}}$ on Dream-7B GSM8K across reverse-step budgets $T$ with $L_{\mathrm{new}}=T$. Ratios below $1$ indicate lower estimated compute for \textsc{LESS}. As $T$ grows, the distribution shifts lower because \textsc{LESS}'s executed step count saturates while the fixed-budget baseline continues to run for all $T$ steps.}
    \label{fig:flops}
\end{figure}

Figure~\ref{fig:flops} shows the corresponding per-example FLOP ratios: as $T$ grows, the ratios concentrate lower as \textsc{LESS}'s executed step count saturates while the fixed-budget sampler continues to execute all $T$ steps. The additional overhead is negligible relative to a Transformer forward pass, since JSD is computed on top-$K$ summaries with $K{=}8$. Wall-clock measurements follow the same trend: on Dream-7B GSM8K, \textsc{LESS} reduces mean latency from $19.45$ to $5.16$ seconds. These results indicate that the efficiency gains come from step-count reduction rather than per-step acceleration.

%% file: sections/07_discussion.tex
\section{Discussion}
\label{sec:disc}

As diffusion LLMs become increasingly practical alternatives to autoregressive generation, inference efficiency becomes central to their real-world applicability. \textsc{LESS} addresses this challenge by committing tokens only when their predictions are confident, temporally persistent, and distributionally stable. Our results show that this stability-gated commitment improves the aggregate accuracy--efficiency trade-off while reducing inference compute, with the most pronounced gains on math and code generation tasks.
\paragraph{Stability is more than confidence.}
The central premise of \textsc{LESS} is that irreversible token commitment should depend on stability at the time of commitment, not solely confidence. A high-confidence predictive distribution can still be unreliable if the top-1 token has recently changed or if probability mass is still moving among plausible alternatives. The ablations in Section~\ref{sec:abl} support this decomposition: confidence provides the main selectivity signal, while top-1 token persistence and inter-step JSD improve robustness by guarding against recent top-1 flips and continuing distributional motion before local predictions have fully settled.

\paragraph{Why inter-step JSD is useful.}
The top-$K$ inter-step JSD provides a practical notion of distributional stability between consecutive token-level predictive distributions. Unlike one-way KL criteria, it does not require choosing a reference step. Its bounded scale gives the JSD threshold a stable numerical range across positions and timesteps, and its symmetry matches the sampler's question: whether two consecutive beliefs are close enough to treat the current prediction as stable, even under changing partially decoded contexts. The top-$K$ projection focuses this comparison on tokens that can plausibly affect the unmasking decision while retaining remaining mass through a residual bin, without requiring full-vocabulary computations.

\paragraph{Complementarity.}
\textsc{LESS} reduces the number of reverse denoising steps and is therefore complementary to methods that reduce per-step cost, such as diffusion-aware KV caching, speculative decoding, or systems-level reuse mechanisms. It is also complementary to training-time acceleration methods that shorten the reverse process through distillation or modified objectives. Because \textsc{LESS} uses only per-position predictive distributions already emitted during decoding, it acts as a drop-in inference-time sampling policy rather than a replacement for model-level or systems-level acceleration, and can be combined with such methods without changing the underlying model.

%% file: sections/08_conclusion.tex
\section{Conclusion}
\label{sec:conclusion}

We introduced \textsc{LESS}, a training-free adaptive sampler for diffusion language models that treats token unmasking as per-position online stopping. \textsc{LESS} commits a masked position only when top-1 confidence, top-1 token persistence, and top-$K$ inter-step Jensen--Shannon divergence jointly indicate stability. Across Dream-7B, LLaDA-8B, and LLaDA-1.5-8B on seven benchmarks, \textsc{LESS} improves the accuracy--efficiency trade-off on average over fixed-budget decoding and strong training-free adaptive samplers, with the largest gains on math and code. Averaged across the three model-family means, it uses $72.1\%$ fewer reverse steps than fixed-budget decoding while also yielding average accuracy gains. Ablations confirm that the three signals provide complementary safeguards, and FLOPs profiling shows that step reductions translate into fewer Transformer forward passes, lower wall-clock time, and lower inference compute. These results support mutual stability as a practical inference-time criterion for robust early token commitment in efficient dLLM sampling.

\section*{Limitations}
\label{sec:limitations}

While \textsc{LESS} provides a training-free mechanism for accelerating diffusion language-model decoding, our study has several limitations. \textsc{LESS} exposes an explicit quality--efficiency trade-off through its stability thresholds, primarily the confidence threshold $c$ and inter-step JSD threshold $d$. More conservative settings, with higher confidence requirements and lower JSD tolerance, reduce the risk of premature commitment but yield more moderate speedups. More permissive settings can accelerate decoding further but may accept unstable positions. In our experiments, this trade-off is resolved by selecting a fixed global configuration shared across the evaluated models and benchmarks. In practice, the optimal operating point may depend on the model, decoding horizon, task type, output length, and tolerance for quality degradation.

%% file: sections/09_appendix.tex
\appendix

\section{Algorithm and Complexity}
\label{app:algorithm}
\label{sec:method:algorithm}

Algorithm~\ref{alg:less} gives the full \textsc{LESS} sampler. For each position that remains masked, the sampler maintains a short top-1-token history $H_i$ and the previous top-$K$ summary $Q_i$ of the predictive distribution. At each reverse step, it forms the eligible set $\mathcal E_t$ using the joint stability rule, selects an unmasking set $\mathcal U_t$ using the unmasking policy from Section~\ref{sec:method:commit}, and applies a forced-unmasking fallback after $\Pi$ consecutive skipped steps.

Given top-$K$ summaries, the coarsened JSD computation is $O(K)$ per masked position. The remaining operations consist of top-1-token-history checks, bounded-window scans, and constant-size state updates. These operations are negligible relative to a Transformer forward pass; the measured speedups therefore come from executing fewer reverse denoising steps rather than from reducing the cost of each step.

\begin{algorithm*}[t]
\caption{\textsc{LESS}: Mutual-Stability Sampling}
\label{alg:less}
\small
\begin{algorithmic}[1]
\REQUIRE model $M$; prompt $x_{\mathrm{prompt}}$; masked sequence $x$; max steps $T$
\REQUIRE thresholds $c,d$; constants $K,P,W,\Pi$
\ENSURE decoded sequence $x$

\STATE Initialize top-1 histories $H_i\gets\emptyset$ and previous top-$K$ summaries $Q_i\gets\emptyset$ for all positions
\STATE $\rho \gets 0$ \hfill{\footnotesize // consecutive skipped steps}

\FOR{$t=T,T-1,\ldots,1$}
    \STATE $\M_t \gets \{i:x_i=\mask\}$
    \IF{$\M_t=\emptyset$}
        \STATE \textbf{return} $x$
    \ENDIF

    \STATE $\{p_{t,i}\}_{i\in\M_t} \gets M(x_{\mathrm{prompt}},x,t)$
    \STATE $\mathcal E_t \gets \{i\in\M_t:\mathrm{Stable}(p_{t,i},H_i,Q_i;c,d,K,P)\}$ \hfill{\footnotesize // eligible positions}

    \STATE $\mathcal U_t \gets \mathrm{FrontierFirst}(\mathcal E_t,\M_t,W)$ \hfill{\footnotesize // selected unmasking set}

    \IF{$\mathcal U_t=\emptyset$ and $\rho<\Pi$}
        \STATE $\rho \gets \rho+1$ \hfill{\footnotesize // defer unmasking}
    \ELSE
        \IF{$\mathcal U_t=\emptyset$}
            \STATE $\mathcal U_t \gets \mathrm{Fallback}(\M_t,\{p_{t,i}\}_{i\in\M_t})$
        \ENDIF
        \STATE $x_{\mathcal U_t}\gets \mathrm{Select}(\{p_{t,i}\}_{i\in\mathcal U_t})$; $\rho\gets0$
    \ENDIF

    \STATE Update $H_i$ and $Q_i$ for all positions that remain masked
\ENDFOR

\STATE \textbf{return} $x$
\end{algorithmic}
\end{algorithm*}

$\mathrm{Stable}$ implements Eq.~\ref{eq:accept}: it computes top-1 confidence, top-1 token persistence, and top-$K$ inter-step JSD, and returns true only if all three tests pass. The set $\mathcal E_t$ contains positions accepted by the joint stability rule, while $\mathcal U_t$ is the set actually unmasked at step $t$. If a position has insufficient top-1-token history or no previous top-$K$ summary, the persistence or JSD test fails by construction. $\mathrm{FrontierFirst}$ returns the frontier position if it is eligible, otherwise the first eligible position within the look-ahead window $W$, and returns $\emptyset$ if none exists. $\mathrm{Fallback}$ applies the underlying sampler's forced-unmasking action, and $\mathrm{Select}$ denotes the model's configured token-selection rule.


\section{\textsc{LESS} Configuration}
\label{app:fixed_constants}

All headline experiments use a single \textsc{LESS} configuration across all model families and benchmarks:
\[
(c,d)=(0.75,0.040).
\]
Here $c$ is the confidence threshold and $d$ is the top-$K$ inter-step JSD threshold. These values define the operating point used throughout the main evaluation. The threshold pair was selected as a stable high-performing operating point on the empirical accuracy--efficiency frontier.

Table~\ref{tab:fixed_constants} summarizes the fixed non-threshold parameters used throughout the paper. We use top-$K$ size $K=8$, persistence $P=2$, frontier look-ahead window $W=16$, and skip budget $\Pi=2$.

\begin{table}[t]
\centering
\small
\setlength{\tabcolsep}{2pt}
\begin{tabular*}{\columnwidth}{@{\extracolsep{\fill}} lcc @{}}
\toprule
Symbol & Description & Value \\
\midrule
$K$ & Top-$K$ inter-step JSD candidate-set size & $8$ \\
$P$ & Top-1 token persistence length & $2$ \\
$W$ & Frontier look-ahead window & $16$ \\
$\Pi$ & Maximum consecutive skipped steps & $2$ \\
\bottomrule
\end{tabular*}
\caption{Fixed non-threshold parameters used by \textsc{LESS} in all experiments.}
\label{tab:fixed_constants}
\end{table}


\section{Evaluation Details}
\label{app:eval-protocol}
\label{app:additional_details}

All evaluations are zero-shot. For each model--benchmark pair, all samplers use the same prompt format, generation length, maximum reverse-step budget, and answer-extraction protocol. We use generation length $L=256$ and reverse-step budget $T=256$ for all benchmarks. For LLaDA-family models, we use semi-autoregressive block size $b=64$ for all benchmarks. GSM8K and MATH are scored by exact-match answer extraction, HumanEval and MBPP by pass@1, and MMLU, HellaSwag, and WinoGrande by accuracy after applying the benchmark-specific answer-extraction protocol. Although the general-knowledge benchmarks require only a short final answer, we keep the same generation length, reverse-step budget, and block size for consistency across tasks and for comparability with the main results in Table~\ref{tab:headline}. Table~\ref{tab:eval-protocol} reports the per-benchmark configuration, including the number of evaluation instances $N$.

\paragraph{Hardware.} All experiments were conducted on \textbf{8$\times$AMD MI210} GPUs, with a total compute allocation of 38 GPU-days.

\begin{table}[t]
\centering
\small
\setlength{\tabcolsep}{4pt}
\begin{tabular*}{\columnwidth}{@{\extracolsep{\fill}} lccccc @{}}
\toprule
Benchmark & $N$ & $L$ & $T$ & LLaDA $b$ & Scoring \\
\midrule
GSM8K      & 1{,}319  & 256 & 256 & 64 & Exact match \\
MATH       & 5{,}000  & 256 & 256 & 64 & Exact match \\
HumanEval  & 164      & 256 & 256 & 64 & Pass@1 \\
MBPP       & 500      & 256 & 256 & 64 & Pass@1 \\
MMLU       & 14{,}042 & 256 & 256 & 64 & Accuracy \\
HellaSwag  & 10{,}042 & 256 & 256 & 64 & Accuracy \\
WinoGrande & 1{,}267  & 256 & 256 & 64 & Accuracy \\
\bottomrule
\end{tabular*}
\caption{Per-benchmark evaluation settings. All evaluations are zero-shot. $N$ is the number of evaluation instances; $L$ is the generation length; $T$ is the maximum reverse-step budget; and $b$ is the LLaDA semi-autoregressive block size. Dream has no block structure. Scoring is exact-match for math, pass@1 for code, and accuracy for general knowledge.}
\label{tab:eval-protocol}
\end{table}

\paragraph{General-knowledge benchmarks.}
MMLU, HellaSwag, and WinoGrande are evaluated in the zero-shot setting on accuracy. For each benchmark and model, the same answer-extraction protocol is applied uniformly across all samplers.

\paragraph{General-knowledge benchmarks.}
MMLU, HellaSwag, and WinoGrande are evaluated in the zero-shot setting on accuracy. For each benchmark and model, the same answer-extraction protocol is applied uniformly across all samplers.


\section{Threshold Robustness}
\label{app:pareto-frontiers}
\label{app:threshold-best}

The main experiments use the single threshold pair $(c,d)=(0.75,0.040)$ across all models and benchmarks. To assess sensitivity to this choice, we sweep $c\in[0.50,0.90]$ and $d\in[0.005,0.080]$ on Dream-7B and LLaDA-8B math and code settings. 

\begin{figure*}[t]
\centering
\includegraphics[width=\textwidth]{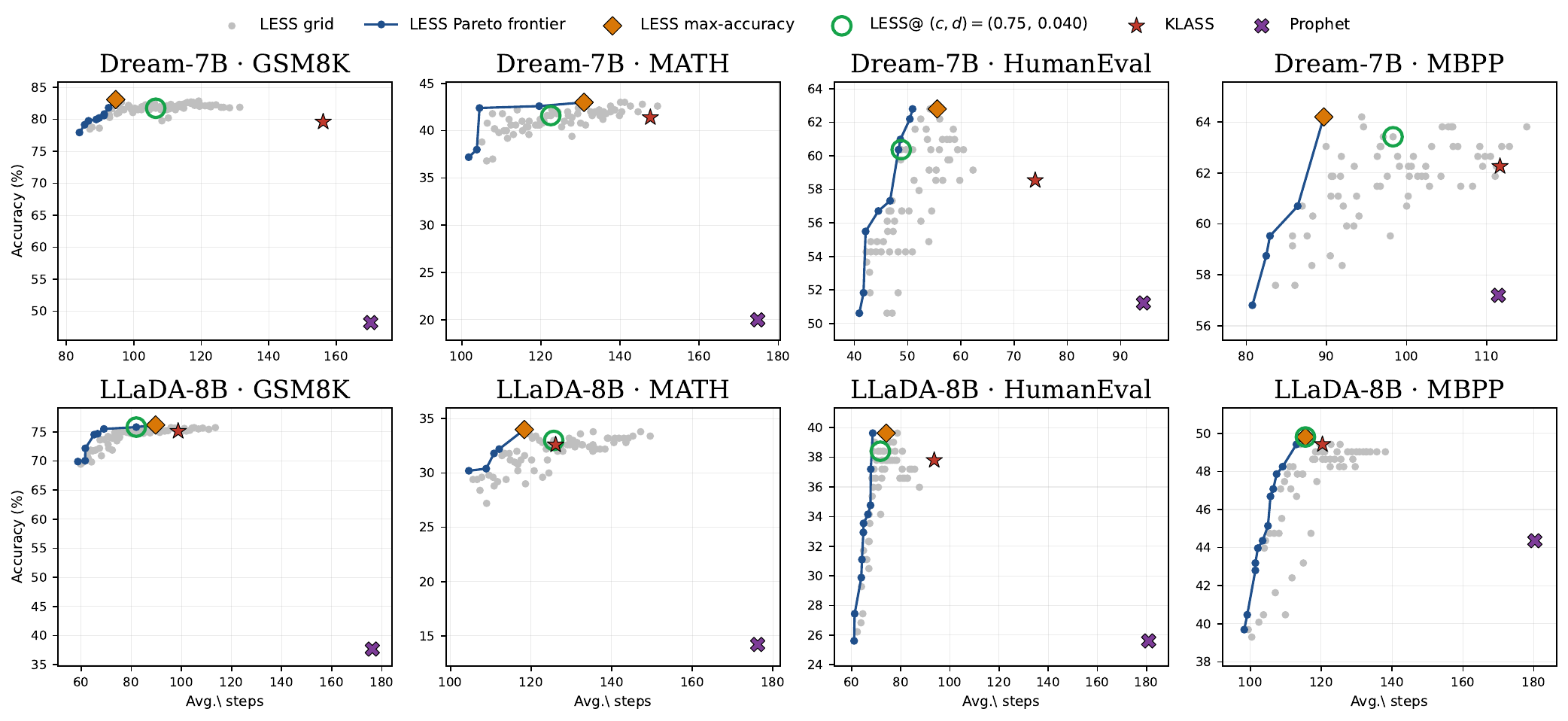}
\caption{\textsc{LESS} threshold sweep over confidence threshold $c$ and inter-step JSD threshold $d$ for Dream-7B and LLaDA-8B on math and code benchmarks. Gray points show swept configurations; the curve shows the empirical Pareto frontier. The sweep varies $c$ from $0.50$ to $0.90$ and $d$ from $0.005$ to $0.080$. Baseline markers are shown for reference.}
\label{fig:pareto-overlay}
\end{figure*}

Figure~\ref{fig:pareto-overlay} shows the resulting accuracy--step trade-offs. The threshold pair lies in a stable high-performing region of the empirical accuracy--efficiency frontier. Per-setting threshold selection can yield additional gains, but the shared configuration captures most of the attainable improvement while avoiding benchmark-specific tuning.

\paragraph{Leave-one-out diagnostic.}
As an additional diagnostic, we evaluate whether the threshold pair is driven by any single model--benchmark setting in the math/code threshold sweep. For each setting, we remove it from the selection pool and choose the threshold pair with the best average accuracy--step trade-off over the remaining settings. This procedure recovers $(c,d)=(0.75,0.040)$ on every split. This indicates that the selected threshold configuration is not driven by a single benchmark or model family, although future model families or substantially different generation regimes may still require calibration.


\section{Full Ablation Results}
\label{app:ablations}

This appendix provides the full ablation tables underlying Section~\ref{sec:abl}. The main text reports the summary conclusions; here we give per-benchmark results for single-signal variants, drop-one variants, unmasking policy, reverse-step budget, token selection, fallback usage, and top-1 token flip rate.

Unless otherwise stated, ablations are run on Dream-7B and LLaDA-8B. We use two thresholding regimes. The single-signal and drop-one ablations use the fixed global configuration from Appendix~\ref{app:fixed_constants}, because these experiments test whether the same criterion can operate across benchmarks without per-task tuning. The reverse-step budget, token-selection, unmasking-policy, and diagnostic analyses use the per-(model, benchmark) $(c,d)$ optimum when stated, in order to isolate the effect of the ablated design choice. Captions specify the thresholding regime used in each table.

Single-signal variants retain only one of the three \textsc{LESS} stability signals. Drop-one variants remove exactly one signal from the full rule. Full \textsc{LESS} denotes the joint top-1-confidence--persistence--JSD rule with the unmasking policy used in the main experiments.

\subsection{Single-Signal Ablations}
\label{app:abl:single}

Each variant retains exactly one of $\{\conf,\persist,\jsd\}$ and disables the other two by neutralizing thresholds: JSD threshold $d\to1.0$, confidence threshold $c\to0$, and persistence length $P\to1$. We additionally include a margin-only variant using the top-1/top-2 logit gap, matching the Prophet-style criterion. All variants are evaluated under the same global-threshold constraint as full \textsc{LESS}: a single threshold is fixed once and applied unchanged across all benchmarks. Specifically, confidence-only uses $c{=}0.75$, JSD-only uses $d{=}0.040$, persistence-only uses $P{=}2$, and margin-only uses $m{=}0.20$, the global threshold with the best mean performance. Full \textsc{LESS} uses the selected global operating point $(c,d){=}(0.75,0.040)$ from Table~\ref{tab:headline}. This protocol evaluates whether each individual signal can serve as a robust task-agnostic commitment criterion under the same cross-benchmark threshold-sharing constraint.
\begin{table*}[t]
\centering
\small
\begin{tabular*}{\textwidth}{@{\extracolsep{\fill}}lcccc@{}}
\toprule
Variant & GSM8K & MATH & HumanEval & MBPP \\
\midrule
\multicolumn{5}{l}{\emph{Dream-7B}} \\
Conf-only ($c{=}0.75$)
  & $81.7 / \phantom{0}84.4$ ($-0.2$) & $43.2 / \phantom{0}98.5$ ($+1.6$) & $59.2 / \phantom{0}40.3$ ($-1.8$) & $62.3 / \phantom{0}84.7$ ($-1.2$) \\
JSD-only ($d{=}0.040$)
  & $30.9 / \phantom{0}44.3$ ($-51.0$) & $10.6 / \phantom{0}45.9$ ($-31.0$) & $15.9 / \phantom{0}24.1$ ($-45.1$) & $10.9 / \phantom{0}35.0$ ($-52.5$) \\
Persistence-only ($P{=}2$)
  & $59.2 / \phantom{0}61.4$ ($-22.7$) & $22.4 / \phantom{0}70.3$ ($-19.2$) & $26.2 / \phantom{0}32.1$ ($-34.8$) & $22.6 / \phantom{0}55.0$ ($-40.9$) \\
Margin-only ($m{=}0.20$)
  & $76.4 / \phantom{0}59.6$ ($-5.5$) & $36.6 / \phantom{0}73.3$ ($-5.0$) & $43.9 / \phantom{0}29.2$ ($-17.1$) & $52.1 / \phantom{0}60.6$ ($-11.3$) \\
\textbf{Full \textsc{LESS}}
  & $81.9 / 106.5$ & $41.6 / 122.5$ & $61.0 / \phantom{0}48.8$ & $63.4 / \phantom{0}98.3$ \\
\midrule
\multicolumn{5}{l}{\emph{LLaDA-8B}} \\
Conf-only ($c{=}0.75$)
  & $75.8 / \phantom{0}56.9$ ($+0.0$) & $33.8 / \phantom{0}94.9$ ($+0.8$) & $36.0 / \phantom{0}51.7$ ($-2.4$) & $48.6 / \phantom{0}86.2$ ($-1.2$) \\
JSD-only ($d{=}0.040$)
  & $\phantom{0}7.0 / \phantom{0}16.3$ ($-68.9$) & $\phantom{0}9.8 / \phantom{0}51.7$ ($-23.2$) & $\phantom{0}1.8 / \phantom{0}26.6$ ($-36.6$) & $\phantom{0}8.2 / \phantom{0}56.4$ ($-41.6$) \\
Persistence-only ($P{=}2$)
  & $\phantom{0}7.4 / \phantom{0}22.3$ ($-68.5$) & $12.8 / \phantom{0}69.5$ ($-20.2$) & $\phantom{0}4.3 / \phantom{0}34.8$ ($-34.1$) & $12.8 / \phantom{0}71.4$ ($-37.0$) \\
Margin-only ($m{=}0.20$)
  & $62.9 / \phantom{0}32.6$ ($-12.9$) & $26.6 / \phantom{0}68.9$ ($-6.4$) & $17.7 / \phantom{0}36.1$ ($-20.7$) & $26.9 / \phantom{0}62.8$ ($-23.0$) \\
\textbf{Full \textsc{LESS}}
  & $75.8 / \phantom{0}82.0$ & $33.0 / 125.6$ & $38.4 / \phantom{0}71.7$ & $49.8 / 115.5$ \\
\bottomrule
\end{tabular*}
\caption{Single-signal ablations on Dream-7B and LLaDA-8B using one fixed global threshold for each variant. Each entry reports accuracy and average reverse steps; $\Delta$ denotes the accuracy difference relative to full \textsc{LESS}. The experiment evaluates whether each individual stability signal can serve as a task-agnostic commitment criterion under the same global-threshold constraint as the full method.}
\label{tab:abl-single}
\end{table*}

Table~\ref{tab:abl-single} shows that no individual signal provides the same reliable operating point as the full mutual-stability rule. Confidence is the strongest standalone signal on average, while JSD-only, persistence-only, and margin-only exhibit substantially weaker aggregate performance under a single global threshold. These results indicate that the stability signals are most effective when used jointly: confidence provides the primary acceptance criterion, while persistence and inter-step distributional stability act as complementary safeguards against premature commitments. Across both model families and all four benchmarks, full \textsc{LESS} provides the most consistent aggregate behavior under the fixed global-threshold setting.

\subsection{Drop-One Ablations}
\label{app:abl:pair}

Each variant disables exactly one of the three stability signals while keeping the remaining criteria at the fixed global configuration. \emph{LESS\,$\setminus$\,persistence} removes the persistence gate and runs with $(c,d){=}(0.75,0.040)$. \emph{LESS\,$\setminus$\,JSD} removes the JSD gate and runs with $c{=}0.75$, $P{=}2$. \emph{LESS\,$\setminus$\,confidence} removes the confidence gate and runs with $d{=}0.040$, $P{=}2$. Full \textsc{LESS} uses the selected global operating point $(c,d){=}(0.75,0.040)$ from Table~\ref{tab:headline}. This experiment evaluates the contribution of each signal to the complete joint stability rule under the same global-threshold constraint.
\begin{table*}[t]
\centering
\small
\begin{tabular*}{\textwidth}{@{\extracolsep{\fill}}lcccc@{}}
\toprule
Variant & GSM8K & MATH & HumanEval & MBPP \\
\midrule
\multicolumn{5}{l}{\emph{Dream-7B}} \\
LESS$\,\setminus$\,persistence
  & $81.0 / \phantom{0}99.4$ ($-0.9$) & $41.4 / 115.1$ ($-0.2$) & $58.5 / \phantom{0}46.6$ ($-2.4$) & $63.0 / \phantom{0}93.3$ ($-0.4$) \\
LESS$\,\setminus$\,JSD
  & $81.9 / \phantom{0}96.4$ ($+0.0$) & $43.0 / 112.1$ ($+1.4$) & $61.0 / \phantom{0}45.3$ ($+0.0$) & $62.3 / \phantom{0}91.0$ ($-1.2$) \\
LESS$\,\setminus$\,confidence
  & $63.2 / \phantom{0}70.2$ ($-18.7$) & $27.4 / \phantom{0}82.8$ ($-14.2$) & $32.9 / \phantom{0}35.0$ ($-28.1$) & $29.6 / \phantom{0}60.0$ ($-33.9$) \\
\textbf{Full \textsc{LESS}}
  & $81.9 / 106.5$ & $41.6 / 122.5$ & $61.0 / \phantom{0}48.8$ & $63.4 / \phantom{0}98.3$ \\
\midrule
\multicolumn{5}{l}{\emph{LLaDA-8B}} \\
LESS$\,\setminus$\,persistence
  & $75.7 / \phantom{0}73.7$ ($-0.1$) & $33.2 / 111.0$ ($+0.2$) & $37.2 / \phantom{0}62.0$ ($-1.2$) & $49.8 / 100.4$ ($+0.0$) \\
LESS$\,\setminus$\,JSD
  & $75.4 / \phantom{0}69.4$ ($-0.5$) & $33.2 / 114.2$ ($+0.2$) & $37.8 / \phantom{0}64.3$ ($-0.6$) & $49.0 / 106.2$ ($-0.8$) \\
LESS$\,\setminus$\,confidence
  & $\phantom{0}9.6 / \phantom{0}25.5$ ($-66.3$) & $16.8 / \phantom{0}79.8$ ($-16.2$) & $\phantom{0}5.5 / \phantom{0}40.2$ ($-32.9$) & $18.7 / \phantom{0}81.2$ ($-31.1$) \\
\textbf{Full \textsc{LESS}}
  & $75.8 / \phantom{0}82.0$ & $33.0 / 125.6$ & $38.4 / \phantom{0}71.7$ & $49.8 / 115.5$ \\
\bottomrule
\end{tabular*}
\caption{Drop-one ablations on Dream-7B and LLaDA-8B using the fixed global configuration. Each entry reports accuracy and average reverse steps; $\Delta$ denotes the accuracy difference relative to full \textsc{LESS}. The experiment evaluates the contribution of each stability signal to the complete mutual-stability rule under a shared global-threshold setting.}
\label{tab:abl-pair}
\end{table*}

Table~\ref{tab:abl-pair} shows that the complete three-signal rule is the most reliable aggregate configuration. Removing confidence leads to large accuracy degradation across both model families, confirming its role as the primary acceptance gate. The two remaining drop-one variants are more competitive, but their aggregate behavior is less stable than the full rule, particularly when evaluated across both model families rather than on a single benchmark in isolation. Persistence and JSD therefore act as complementary stabilizers: each helps reduce the risk of accepting tokens whose predictions have not yet settled, and their combination yields the strongest global operating configuration. Together with the single-signal results, these ablations support the use of \textsc{LESS} as a joint stability criterion rather than a standalone confidence, margin, persistence, or distributional-drift rule.

\subsection{Unmasking Policy}
\label{app:abl:commit}

To isolate the effect of the unmasking policy, we compare the frontier-first policy used by full \textsc{LESS} against a parallel-unmasking variant that commits all eligible positions at each reverse step. The stability criteria are kept fixed at the per-(model, benchmark) $(c,d)$ optimum. This ablation tests whether the ordering of accepted positions affects accuracy and reverse-step usage once the same joint stability rule has identified eligible positions.

\begin{table*}[t]
\centering
\small
\begin{tabular*}{\textwidth}{@{\extracolsep{\fill}}lcccc@{}}
\toprule
Variant & GSM8K & MATH & HumanEval & MBPP \\
\midrule
\multicolumn{5}{l}{\emph{Dream-7B}} \\
Parallel unmasking
  & $78.32 / 109.6$ & $42.80 / 112.4$ & $56.71 / 44.2$ & $64.59 / 104.5$ \\
$\Delta$ vs.\ full \textsc{LESS}
  & $-4.77$ / $+15.0$ & $-0.20$ / $-18.5$ & $-6.09$ / $-6.7$ & $+0.39$ / $+14.8$ \\
Full \textsc{LESS} (frontier)
  & $83.09 / 94.6$ & $43.00 / 130.9$ & $62.80 / 50.9$ & $64.20 / 89.7$ \\
\midrule
\multicolumn{5}{l}{\emph{LLaDA-8B}} \\
Parallel unmasking
  & $76.19 / \phantom{0}89.7$ & $34.00 / 118.3$ & $39.63 / 78.6$ & $49.81 / 114.6$ \\
$\Delta$ vs.\ full \textsc{LESS}
  & $\phantom{-}0.00$ / $\phantom{-1}0.0$ & $\phantom{-}0.00$ / $\phantom{-1}0.0$ & $\phantom{-}0.00$ / $+10.0$ & $\phantom{-}0.00$ / $\phantom{-1}0.0$ \\
Full \textsc{LESS} (frontier)
  & $76.19 / 89.7$ & $34.00 / 118.3$ & $39.63 / 68.6$ & $49.81 / 114.6$ \\
\bottomrule
\end{tabular*}
\caption{Unmasking-policy ablation comparing parallel unmasking against the frontier-first policy used by full \textsc{LESS} at the per-(model, benchmark) $(c,d)$ optimum. Method rows report accuracy and average executed reverse steps; $\Delta$ rows report differences relative to full \textsc{LESS}. On LLaDA-8B, the two policies produce identical accuracy because the released semi-autoregressive blockwise reverse process already unmasks accepted positions in parallel within the active block.}
\label{tab:abl-commit}
\end{table*}

Table~\ref{tab:abl-commit} isolates the effect of the unmasking policy after eligibility has been determined by the same stability rule. On Dream-7B, the frontier-first policy improves aggregate accuracy relative to parallel unmasking, consistent with the interpretation that committing many eligible positions simultaneously can perturb subsequent denoising states before local context has fully settled. On LLaDA-8B, accuracy is unchanged because the released semi-autoregressive blockwise reverse process already commits accepted positions in parallel within the active block. These results indicate that the frontier-first policy is most relevant in the full-sequence setting, where commitment order directly affects subsequent denoising states.

\subsection{Reverse-Step Budget}
\label{app:abl:budget}

We vary $T\in\{64,128,256,512\}$ at the per-(model, benchmark) $(c,d)$ optimum and report accuracy and average executed reverse steps.

\begin{table*}[t]
\centering
\small
\begin{tabular*}{\textwidth}{@{\extracolsep{\fill}}lcccc@{}}
\toprule
$T$ & GSM8K & MATH & HumanEval & MBPP \\
\midrule
\multicolumn{5}{l}{\emph{Dream-7B}} \\
$64$  & $18.20 / 62.6$  & $\phantom{0}2.80 / 63.8$  & $56.10 / 40.1$ & $62.65 / 60.0$ \\
$128$ & $79.45 / 93.2$  & $29.20 / 117.6$ & $62.80 / 49.0$ & $63.42 / 86.0$ \\
$256$ & $83.09 / 94.6$ & $43.00 / 130.9$ & $62.80 / 50.9$ & $64.20 / 89.7$ \\
$512$ & $83.32 / 94.7$  & $43.00 / 130.9$ & $62.80 / 50.9$ & $64.20 / 89.7$ \\
\midrule
\multicolumn{5}{l}{\emph{LLaDA-8B}} \\
$64$  & $67.70 / 54.6$ & $21.20 / \phantom{0}60.1$ & $12.20 / 42.9$ & $\phantom{0}6.61 / \phantom{0}58.5$ \\
$128$ & $74.98 / 75.1$ & $31.00 / \phantom{0}93.3$ & $29.27 / 60.0$ & $26.46 / \phantom{0}88.6$ \\
$256$ & $76.19 / 89.7$ & $34.00 / 118.3$ & $39.63 / 68.6$ & $49.81 / 114.6$ \\
$512$ & $76.19 / 89.7$ & $34.00 / 118.3$ & $39.63 / 68.6$ & $49.81 / 114.6$ \\
\bottomrule
\end{tabular*}
\caption{Reverse-step budget ablation on Dream-7B and LLaDA-8B at the per-(model, benchmark) $(c,d)$ optimum. Each entry reports accuracy and average executed reverse steps.}
\label{tab:abl-budget}
\end{table*}

Table~\ref{tab:abl-budget} shows that the rule reaches its operating point at $T=256$ for the evaluated settings. Increasing the budget to $T=512$ produces no material aggregate improvement, while reducing it to $T=128$ degrades harder settings, especially LLaDA and Dream-7B MATH. At $T=64$, Dream-7B GSM8K and MATH degrade sharply, indicating that the sampler cannot defer commitment long enough when the reverse budget ends before the predictive distributions stabilize.

\subsection{Token Selection}
\label{app:abl:stoch}

The eligibility signals in \textsc{LESS} are computed from the model predictive distribution $p_{t,i}$ before token selection. We therefore vary the token-selection rule while keeping the eligibility rule fixed, comparing argmax selection with two nucleus-sampling configurations for both model families.

\begin{table*}[t]
\centering
\small
\begin{tabular*}{\textwidth}{@{\extracolsep{\fill}}lcccc@{}}
\toprule
Selection & GSM8K & MATH & HumanEval & MBPP \\
\midrule
\multicolumn{5}{l}{\emph{Dream-7B}} \\
Argmax
  & $83.09 / 94.6$ & $43.00 / 130.9$ & $62.80 / 50.9$ & $64.20 / 89.7$ \\
Nucleus $\tau{=}0.5,\,p{=}0.95$
  & $79.68 / 95.2$ & $42.40 / 131.1$ & $62.80 / 50.9$ & $62.65 / 89.8$ \\
Nucleus $\tau{=}0.7,\,p{=}0.90$
  & $77.86 / 96.1$ & $43.40 / 131.3$ & $61.59 / 51.5$ & $54.86 / 88.7$ \\
\midrule
\multicolumn{5}{l}{\emph{LLaDA-8B}} \\
Argmax
  & $76.19 / 89.7$ & $34.00 / 118.3$ & $39.63 / 68.6$ & $49.81 / 114.6$ \\
Nucleus $\tau{=}0.5,\,p{=}0.95$
  & $54.51 / 46.4$ & $24.60 / 102.2$ & $12.80 / 55.8$ & $26.46 / \phantom{0}95.5$ \\
Nucleus $\tau{=}0.7,\,p{=}0.90$
  & $60.20 / 53.4$ & $26.80 / 114.7$ & $20.12 / 64.4$ & $31.91 / 104.7$ \\
\bottomrule
\end{tabular*}
\caption{Token-selection ablation on Dream-7B and LLaDA-8B at the per-(model, benchmark) $(c,d)$ optimum. Each entry reports accuracy and average reverse steps.}
\label{tab:abl-stoch}
\end{table*}

Table~\ref{tab:abl-stoch} shows that accuracy changes mainly reflect each model's sensitivity to token selection rather than changes to the eligibility criterion. Dream-7B is less sensitive to the tested nucleus-sampling variants than LLaDA-8B, which shows substantial degradation across all four benchmarks. The eligibility criterion itself is independent of the token-selection rule at each reverse step, since it operates on $p_{t,i}$ before any token is selected.

\subsection{Fallback Usage and Top-1 Token Stability}
\label{app:abl:fallback}

We audit Dream-7B on 50 examples per benchmark at the per-setting tuned configuration. For each reverse step, we log whether the fallback is invoked, the number of positions unmasked, and the top-1 token at each position. The fallback is invoked only when no position is accepted by the joint stability rule at a given step; the forced-unmasking action then unmasks the frontier position.

Across GSM8K, MATH, HumanEval, and MBPP, the fallback is invoked on $2.1\%$, $3.9\%$, $8.9\%$, and $2.5\%$ of reverse steps, respectively. Thus, most Dream-7B reverse steps are governed directly by the joint stability rule, and \textsc{LESS} acts primarily as a stability-gated stopping rule in the full-sequence setting. The per-step top-1 token flip rate on Dream-7B is highest early in decoding and falls to at most $5\%$ in the final quarter of the trajectory across all audited benchmarks. This supports the temporal-stability assumption underlying \textsc{LESS}: per-position predictive distributions tend to stabilize over reverse time, but early commitments can occur before that stabilization is complete.


\section{Compute and Latency}
\label{app:compute}
\label{app:wallclock}
\label{app:flops_details}

This appendix section expands the compute analysis in Section~\ref{subsec:flops}. All compared samplers use the same Transformer forward pass at a given sequence length, so differences in total compute are driven primarily by the number of reverse denoising steps executed. \textsc{LESS} reduces this count by committing stable positions earlier and terminating once all positions are unmasked.

\paragraph{Wall-clock measurements.}
We measure single-GPU wall-clock decoding time on Dream-7B GSM8K. All methods use $T=256$ and $L=256$, with synchronized accelerator timing and two warmup runs per method.

\begin{table}[t]
\centering
\small
\begin{tabular*}{\columnwidth}{@{\extracolsep{\fill}} lcc @{}}
\toprule
Method & Seconds/sample & Speedup vs. Base \\
\midrule
Base & $19.45 \pm 0.87$ & $1.00\times$ \\
Prophet & $7.20 \pm 3.09$ & $2.70\times$ \\
KLASS & $7.62 \pm 2.96$ & $2.55\times$ \\
\textsc{LESS} & $\mathbf{5.16 \pm 2.00}$ & $\mathbf{3.77\times}$ \\
\bottomrule
\end{tabular*}
\caption{Single-GPU wall-clock decoding time on Dream-7B GSM8K.}
\label{tab:wallclock}
\end{table}

Table~\ref{tab:wallclock} shows that the wall-clock measurements follow the same trend as reverse-step counts. Since \textsc{LESS} does not modify the Transformer forward pass, the speedup comes from fewer reverse steps. Sampling overhead is below measurement resolution relative to a full model forward pass.

\paragraph{Forward-pass accounting.}
For a sequence of length $N$, we approximate the cost of one Dream-7B Transformer forward pass as
\begin{equation}
\begin{split}
F(N) \;\approx{}&
n_{\mathrm{layer}} \cdot 2(2 d_{\mathrm{h}}^2 + 2 d_{\mathrm{h}} d_{\mathrm{kv}} + 3 d_{\mathrm{h}} d_{\mathrm{ffn}})\,N \\
&{}+ n_{\mathrm{layer}} \cdot 4\, d_{\mathrm{h}}\, N^2
+ 2\, d_{\mathrm{h}}\, |\V|\, N .
\end{split}
\label{eq:flops_F_N}
\end{equation}
The first term sums per-layer linear projections, the second term is attention, and the third term is the language-model head. We use Dream-7B's configuration: $n_{\mathrm{layer}}=28$, $d_{\mathrm{h}}=3584$, $d_{\mathrm{kv}}=4\cdot128=512$, $d_{\mathrm{ffn}}=18944$, and $|\V|=152064$. At our generation lengths, the linear terms dominate and the attention quadratic contributes at most $5\%$ of $F(N)$ across the sweep \citep{kaplan2020scaling,narayanan2021efficient}.

\paragraph{FLOPs profiling protocol.}
We profile Dream-7B on GSM8K using $T\in\{32,64,128,256\}$ with $L_{\mathrm{new}}=T$. Each sampler records the number of Transformer forward calls executed per example. FLOPs are estimated from measured per-forward costs and counted model forward evaluations. We report the mean of per-example FLOP speedup ratios, $\mathbb E[\mathrm{FLOPs}_{\mathrm{Base}}/\mathrm{FLOPs}_{\textsc{LESS}}]$. This quantity is not generally equal to $T/\mathbb E[\#\mathrm{forwards}]$, because examples terminate after different numbers of forward calls and the table averages ratios rather than taking the ratio of averages.

\begin{table}[t]
\centering
\small
\setlength{\tabcolsep}{6pt}
\begin{tabular*}{\columnwidth}{@{\extracolsep{\fill}} rcc @{}}
\toprule
$T$ & Speedup $\uparrow$ & \textsc{LESS} forwards $\downarrow$ \\
\midrule
 32  & $1.40 \pm 0.57$ & $25.14 \pm \phantom{0}5.66$ \\
 64  & $1.46 \pm 0.45$ & $45.41 \pm \phantom{0}6.26$ \\
128  & $2.17 \pm 0.89$ & $65.65 \pm 18.05$ \\
256  & $5.01 \pm 3.08$ & $64.48 \pm 30.52$ \\
\bottomrule
\end{tabular*}
\caption{FLOPs profiling on Dream-7B GSM8K ($n{=}1{,}319$) across reverse-step budgets $T$ with $L_{\mathrm{new}}{=}T$. We report mean $\pm$ standard deviation over examples. Speedup is the mean per-example estimated ratio $\mathrm{FLOPs}_{\textsc{Base}}/\mathrm{FLOPs}_{\textsc{LESS}}$; \textsc{LESS} forwards denotes the number of model forward evaluations executed by \textsc{LESS}. Because speedup is averaged per example, it need not equal $T$ divided by the mean \textsc{LESS} forward count.}
\label{tab:flops_profile_gsm8k}
\end{table}

Table~\ref{tab:flops_profile_gsm8k} shows that, at large budgets, \textsc{LESS}'s forward-pass count saturates near $65$ calls on average, while the fixed-budget sampler continues to execute all $T$ steps. This indicates that \textsc{LESS}'s efficiency gains come from step-count reduction rather than per-step model acceleration.


\section{Qualitative Examples}
\label{app:qualitative}

Table~\ref{tab:headline} reports aggregate accuracy and reverse-step counts. This appendix provides four qualitative Dream-7B examples to illustrate representative output-level behavior: two Python code-synthesis prompts in Tables~\ref{tab:qual:04_write_a_python_program_to_reve} and~\ref{tab:qual:01_write_a_python_program_to_calc}, and two elementary arithmetic prompts in Tables~\ref{tab:qual:02_what_is_the_sum_of_the_first_1} and~\ref{tab:qual:03_solve_the_equation_3x_5_20}. These examples are illustrative; the quantitative conclusions are based on the benchmark results and ablations reported in the main text and appendix.

\paragraph{Setup.}
We compare Base, Prophet, KLASS, and \textsc{LESS} using the same Dream-7B checkpoint, temperature $0.2$, nucleus parameter $p=0.95$, maximum generation length $256$, and reverse-step budget $T=256$. Base executes the full reverse process; Prophet uses its published default phase thresholds; KLASS uses its released HumanEval configuration; and \textsc{LESS} uses the fixed global configuration $(c,d)=(0.75,0.040)$ from Appendix~\ref{app:fixed_constants}. For each sampler, we report generated text, executed reverse steps, and end-to-end wall-clock latency under the hardware setup in Appendix~\ref{app:compute}.

Each qualitative table includes the prompt, a compact reference answer, sampler output, executed reverse steps, latency, and a correctness marker. The code examples illustrate structured-output generation, where correctness depends on syntax, control flow, and boundary conventions. The arithmetic examples illustrate a simpler regime in which the final answer often stabilizes before the full reverse budget is exhausted, allowing \textsc{LESS} to terminate earlier once the relevant positions satisfy the mutual-stability criterion. These examples are consistent with the aggregate trends in Table~\ref{tab:headline}, but serve only as qualitative illustrations.
\input{sections/qualitative_table}

%% file: sections/qualitative_table.tex
\begin{table*}[t]
\centering
\arrayrulecolor{QualRule}
\renewcommand{\arraystretch}{0.95}
\setlength{\extrarowheight}{0pt}
\begin{tabular}{|>{\columncolor{QualHeaderBG}\raggedright\arraybackslash}p{2.6cm}|p{\dimexpr\textwidth-2.6cm-4\tabcolsep-3\arrayrulewidth\relax}|}
\hline
\rowcolor{QualHeaderBG}
\textbf{Question} &
{\small\itshape Write a Python program to reverse a string.} \\
\hline

\rowcolor{QualHeaderBG}
\textbf{Answer} &
{\small A function returning the input string with characters in reverse order, e.g.\ \texttt{reverse("abc")="cba"}.} \\
\hline

\makecell[l]{\textbf{Base}\,\qualbad\\[1pt]\qualmeta{256 steps}\\\qualmeta{15.15\,s}} &
{\vspace{-22pt}\small Here is the code:}
\begin{lstlisting}[style=qualpython]
def reverse(string):
    reverse_string = string

input_string = "Hello, World!"
reversed_string = reverse(input_string)
print(reversed_string)
\end{lstlisting} \\
\hline

\makecell[l]{\textbf{Prophet}\,\qualok\\[1pt]\qualmeta{132 steps}\\\qualmeta{9.14\,s}} &
{\vspace{-22pt}\small Here is a simple Python program to reverse a string:}
\begin{lstlisting}[style=qualpython]
def reverse_string(input_string):
    return input_string[::-1]

input_string = input("Enter a string: ")
print(reverse_string(input_string))
\end{lstlisting} \\
\hline

\makecell[l]{\textbf{KLASS}\,\qualok\\[1pt]\qualmeta{134 steps}\\\qualmeta{9.80\,s}} &
{\vspace{-22pt}\small Here is a simple Python program to reverse a string:}
\begin{lstlisting}[style=qualpython]
def reverse_string(input_string):
    return input_string[::-1]

input_string = input("Enter a string: ")
print(reverse_string(input_string))
\end{lstlisting} \\
\hline

\makecell[l]{\textbf{\textsc{LESS} (ours)}\,\qualok\\[1pt]\qualmeta{117 steps}\\\qualmeta{8.13\,s}} &
{\vspace{-22pt}\small Here is a simple Python program to reverse a string:}
\begin{lstlisting}[style=qualpython]
def reverse_string(input_string):
    return input_string[::-1]

input_string = "Hello, World!"
print(reverse_string(input_string))
\end{lstlisting} \\
\hline

\end{tabular}
\arrayrulecolor{black}
\caption{Generated solutions on Dream-7B for \emph{Write a Python program to reverse a string.}}
\label{tab:qual:04_write_a_python_program_to_reve}
\end{table*}


\begin{table*}[!htbp]
\centering
\arrayrulecolor{QualRule}
\renewcommand{\arraystretch}{0.95}
\setlength{\extrarowheight}{0pt}

\begin{tabular}{|>{\columncolor{QualHeaderBG}\raggedright\arraybackslash}p{2.6cm}|p{\dimexpr\textwidth-2.6cm-4\tabcolsep-3\arrayrulewidth\relax}|}
\hline
\rowcolor{QualHeaderBG}
\textbf{Question} &
{\small\itshape Write a python program to calculate the Fibonacci of a given number.} \\
\hline

\rowcolor{QualHeaderBG}
\textbf{Answer} &
{\small A function returning the $n$-th Fibonacci number with base cases $\mathrm{fib}(0){=}0$, $\mathrm{fib}(1){=}1$ and recurrence $\mathrm{fib}(n){=}\mathrm{fib}(n{-}1){+}\mathrm{fib}(n{-}2)$.} \\
\hline

\makecell[l]{\textbf{Base}\,\qualok\\[1pt]\qualmeta{256 steps}\\\qualmeta{15.35\,s}} &
\vspace{-25pt}
\begin{lstlisting}[style=qualpython]
def Fibonacci(n):
    if n == 0:
        return 0
    elif n == 1:
        return 1
    elif n > 1:
        return Fibonacci(n-1) + Fibonacci(n-2)
\end{lstlisting} \\
\hline

\makecell[l]{\textbf{Prophet}\,\qualbad\\[1pt]\qualmeta{250 steps}\\\qualmeta{16.18\,s}} &
{\vspace{-22pt}\small Here is a simple Python program to calculate the Fibonacci of a given number:}
\begin{lstlisting}[style=qualpython]
def fibonacci(n):
    if n <= 0:
        return "Input should be a positive integer."
    elif n == 1:
        return 0
    elif n == 2:
        return 1
    else:
        return fibonacci(n-1) + fibonacci(n-2)

num = int(input("Enter a number: "))
print(fibonacci(num))
\end{lstlisting} \\
\hline

\makecell[l]{\textbf{KLASS}\,\qualok\\[1pt]\qualmeta{219 steps}\\\qualmeta{15.51\,s}} &
{\vspace{-22pt}\small Here is a simple Python program to calculate the Fibonacci of a given number:}
\begin{lstlisting}[style=qualpython]
def fibonacci(n):
    if n <= 0:
        return 0
    elif n == 1:
        return 1
    else:
        return fibonacci(n-1) + fibonacci(n-2)

n = int(input("Enter a number: "))
print("Fibonacci of", n, "is", fibonacci(n))
\end{lstlisting} \\
\hline

\makecell[l]{\textbf{\textsc{LESS} (ours)}\,\qualok\\[1pt]\qualmeta{119 steps}\\\qualmeta{8.09\,s}} &
{\vspace{-22pt}\small Here is a simple Python program that calculates the Fibonacci of a given number using recursion:}
\begin{lstlisting}[style=qualpython]
def fibonacci(n):
    if n <= 0:
        return 0
    elif n == 1:
        return 1
    else:
        return fibonacci(n-1) + fibonacci(n-2)

# Test the function
n = 10
print(f"Fibonacci of {n} is {fibonacci(n)}")
\end{lstlisting} \\
\hline

\end{tabular}
\arrayrulecolor{black}
\caption{Generated solutions on Dream-7B for \emph{Write a python program to calculate the Fibonacci of a given number.}}
\label{tab:qual:01_write_a_python_program_to_calc}
\end{table*}


\begin{table*}[!htbp]
\centering
\arrayrulecolor{QualRule}
\renewcommand{\arraystretch}{0.95}
\setlength{\extrarowheight}{0pt}

\begin{tabular}{|>{\columncolor{QualHeaderBG}\raggedright\arraybackslash}p{2.6cm}|p{\dimexpr\textwidth-2.6cm-4\tabcolsep-3\arrayrulewidth\relax}|}
\hline
\rowcolor{QualHeaderBG}
\textbf{Question} &
{\small\itshape What is the sum of the first 100 natural numbers?} \\
\hline

\rowcolor{QualHeaderBG}
\textbf{Answer} &
{\small $\sum_{i=1}^{100} i \;=\; \frac{100 \cdot 101}{2} \;=\; 5050$} \\
\hline

\makecell[l]{\textbf{Base}\,\qualok\\[1pt]\qualmeta{255 steps}\\\qualmeta{16.23\,s}} &
{\vspace{-22pt}\small
The sum of the first $n$ natural numbers is given by the formula $n(n+1)/2$.\par
In this case, $n = 100$, so the sum is $100(101)/2 = 5050$.\par
The answer is 5050.
} \\
\hline

\makecell[l]{\textbf{Prophet}\,\qualok\\[1pt]\qualmeta{80 steps}\\\qualmeta{5.99\,s}} &
{\vspace{-22pt}\small
The sum of the first $n$ natural numbers is given by the formula:\par
$\mathrm{Sum} = n(n + 1)/2$\par
In this case, $n = 100$, so:\par
$\mathrm{Sum} = 100(100 + 1)/2$\par
$\mathrm{Sum} = 100(101)/2$\par
$\mathrm{Sum} = 5050$\par
Therefore, the sum of the first 100 natural numbers is 5050.
} \\
\hline

\makecell[l]{\textbf{KLASS}\,\qualok\\[1pt]\qualmeta{74 steps}\\\qualmeta{5.63\,s}} &
{\vspace{-22pt}\small
The sum of the first $n$ natural numbers is given by the formula:\par
$\mathrm{Sum} = n(n + 1) / 2$\par
In this case, $n = 100$, so:\par
$\mathrm{Sum} = 100(100 + 1) / 2$\par
$\mathrm{Sum} = 100 \cdot 101 / 2$\par
$\mathrm{Sum} = 5050$\par
Therefore, the sum of the first 100 natural numbers is 5050.
} \\
\hline

\makecell[l]{\textbf{\textsc{LESS} (ours)}\,\qualok\\[1pt]\qualmeta{24 steps}\\\qualmeta{1.89\,s}} &
{\vspace{-22pt}\small
The sum of the first 100 natural numbers is $\frac{100 \cdot 101}{2} = \boxed{5050}$.\par
The answer is: 5050.
} \\
\hline

\end{tabular}
\arrayrulecolor{black}
\caption{Generated solutions on Dream-7B for \emph{What is the sum of the first 100 natural numbers?}}
\label{tab:qual:02_what_is_the_sum_of_the_first_1}
\end{table*}


\begin{table*}[!htbp]
\centering
\arrayrulecolor{QualRule}
\renewcommand{\arraystretch}{0.95}
\setlength{\extrarowheight}{0pt}

\begin{tabular}{|>{\columncolor{QualHeaderBG}\raggedright\arraybackslash}p{2.6cm}|p{\dimexpr\textwidth-2.6cm-4\tabcolsep-3\arrayrulewidth\relax}|}
\hline
\rowcolor{QualHeaderBG}
\textbf{Question} &
{\small\itshape Solve the equation: 3x + 5 = 20.} \\
\hline

\rowcolor{QualHeaderBG}
\textbf{Answer} &
{\small $x = 5$} \\
\hline

\makecell[l]{\textbf{Base}\,\qualok\\[1pt]\qualmeta{256 steps}\\\qualmeta{16.32\,s}} &
{\vspace{-22pt}\small $x = 5$} \\
\hline

\makecell[l]{\textbf{Prophet}\,\qualok\\[1pt]\qualmeta{140 steps}\\\qualmeta{10.31\,s}} &
{\vspace{-22pt}\small
To solve the equation $3x + 5 = 20$, we need to isolate the variable $x$ on one side of the equation.\par
First, subtract 5 from both sides:\par
$3x + 5 - 5 = 20 - 5$\par
$3x = 15$\par
Next, divide both sides by 3:\par
$3x / 3 = 15 / 3$\par
$x = 5$\par
So, the solution to the equation $3x + 5 = 20$ is $x = 5$.
} \\
\hline

\makecell[l]{\textbf{KLASS}\,\qualok\\[1pt]\qualmeta{103 steps}\\\qualmeta{7.68\,s}} &
{\vspace{-22pt}\small
To solve the equation $3x + 5 = 20$, we need to isolate the variable $x$ on one side of the equation.\par
First, subtract 5 from both sides to get $3x = 15$.\par
Next, divide both sides by 3 to solve for $x$: $x = 5$.\par
So, the solution to the equation $3x + 5 = 20$ is $x = 5$.
} \\
\hline

\makecell[l]{\textbf{\textsc{LESS} (ours)}\,\qualok\\[1pt]\qualmeta{56 steps}\\\qualmeta{4.29\,s}} &
{\vspace{-22pt}\small
To solve the equation $3x + 5 = 20$, we need to isolate the variable $x$.\par
First, subtract 5 from both sides:\par
$3x + 5 - 5 = 20 - 5$\par
$3x = 15$\par
Now, divide both sides by 3:\par
$3x / 3 = 15 / 3$\par
$x = 5$\par
So, the solution to the equation $3x + 5 = 20$ is $x = 5$.
} \\
\hline

\end{tabular}
\arrayrulecolor{black}
\caption{Generated solutions on Dream-7B for \emph{Solve the equation: 3x + 5 = 20.}}
\label{tab:qual:03_solve_the_equation_3x_5_20}
\end{table*}